 \documentclass[final,5p,times,twocolumn,authoryear]{elsarticle}

\usepackage{amssymb}
\usepackage{float}
\usepackage{pgfplots}
\usepackage{tikz}
\usetikzlibrary{patterns}
\usepackage{lipsum}
\usepackage{hyperref}
\hypersetup{
    colorlinks=true,
    linkcolor=blue,
    citecolor=blue
    }
\usepackage{amsthm}
\usepackage{amsfonts}
\usepackage{amsmath}

\journal{Computers in Biology and Medicine}

\begin{document}

\begin{frontmatter}



\title{Reducing Intraspecies and Interspecies Covariate Shift in Traumatic Brain Injury EEG of Humans and Mice Using Transfer Euclidean Alignment}



\author[first]{Manoj Vishwanath}
\author[first,second]{Steven Cao}
\author[first,second]{Nikil Dutt}
\author[first,second]{Amir M. Rahmani}
\author[third,fourth]{Miranda M. Lim}
\author[first,second,fifth]{Hung Cao}
\affiliation[first]{organization={Department of Computer Science, University of California },
            city={Irvine},
            postcode={92697}, 
            state={CA},
            country={USA}}
\affiliation[second]{organization={Department of Electrical Engineering and Computer Science, University of California },
            city={Irvine},
            postcode={92697}, 
            state={CA},
            country={USA}}

\affiliation[third]{organization={VA Portland Health Care System },
            city={Portland},
            postcode={97239}, 
            state={OR},
            country={USA}}
\affiliation[fourth]{organization={Oregon Health \& Science University },
            city={Portland},
            postcode={97239}, 
            state={OR},
            country={USA}}
\affiliation[fifth]{organization={Department of Biomedical Engineering, University of California },
            city={Irvine},
            postcode={92697}, 
            state={CA},
            country={USA}}

\begin{abstract}
While analytics of sleep electroencephalography (EEG) holds certain advantages over other methods in clinical applications, high variability across subjects poses a significant challenge when it comes to deploying machine learning models for classification tasks in the real world. In such instances, machine learning models that exhibit exceptional performance on a specific dataset may not necessarily demonstrate similar proficiency when applied to a distinct dataset for the same task. The scarcity of high-quality biomedical data further compounds this challenge, making it difficult to evaluate the model's generality comprehensively. In this paper, we introduce Transfer Euclidean Alignment - a transfer learning technique to tackle the problem of the dearth of human biomedical data for training deep learning models. We tested the robustness of this transfer learning technique on various rule-based classical machine learning models as well as the EEGNet-based deep learning model by evaluating on different datasets, including human and mouse data in a binary classification task of detecting individuals with versus without traumatic brain injury (TBI). By demonstrating notable improvements with an average increase of 14.42\% for intraspecies datasets and 5.53\% for interspecies datasets, our findings underscore the importance of the use of transfer learning to improve the performance of machine learning and deep learning models when using diverse datasets for training.

\end{abstract}




\end{frontmatter}




\section{Introduction}
\label{introduction}

Electroencephalogram (EEG) is a widely used non-invasive technique for assessing the electrical activity of the brain. Electrical signals generated by the synchronous activity of a population of neurons are picked up by the electrodes placed on the scalp which are then amplified and recorded, providing a means of understanding the human brain \cite{nunez2006electric}. Quantitative EEG (QEEG) is a method of analyzing EEG data that involves the computation of its various measures, such as power spectral density, coherence, and phase synchrony \cite{prichep2005use}. EEG has been used in a wide range of clinical applications, including the diagnosis and monitoring of epilepsy \cite{acharya2013automated,noachtar2009role}, sleep disorders, stroke \cite{wu2015connectivity}, traumatic brain injury (TBI) \cite{rapp2015traumatic,vishwanath2021detection}, and Alzheimer's disease \cite{jeong2004eeg}, in addition to the study of cognitive processes such as attention \cite{liu2013recognizing}, memory \cite{klimesch1999eeg}, language \cite{weiss2003contribution}, and general intelligence \cite{thatcher2016intelligence}, as it can reveal anomalies that are indicative of brain dysfunction. Despite the benefits of EEG for real-time monitoring, such as affordability and negligible dependence on bulky equipment and highly trained clinicians, EEG poses certain challenges, such as its non-stationary characteristics and low signal-to-noise ratio \cite{nunez2006electric}. Moreover, the high variability of the EEG signal across sessions and individuals can pose difficulties in generalizing results \cite{xu2020cross}.

Despite constant efforts to build models using advanced machine learning (ML) and deep learning (DL) algorithms that account for cross-subject variability, the heterogeneous nature of EEG leads to inconsistent results, hindering model transportability. In most cases, models that perform extremely well on one dataset may not necessarily perform well on another dataset carrying out the same task due to overfitting of the model or inherent differences present amongst the datasets \cite{kamrud2021effects}. ML algorithms rely on a crucial assumption that both the training and testing datasets of a classifier share the same feature space and follow the same distribution. However, in real-world scenarios, this assumption may not always hold, which often leads to challenges in practical implementations of the ML model \cite{pan2010survey}. To address this issue, traditional ML approaches are often carried out using a vast amount of training data to capture the maximum variability. However, it is extremely difficult to collect a sufficient amount of high-quality data in the biomedical domain due to concerns of privacy and confidentiality, lack of standardization, ethical issues, and incentives for data sharing. Although validation sets, ensemble models, and data augmentation are commonly used solutions for addressing the problem of limited training data, they all have certain limitations and assumptions. The use of a validation set for parameter tuning can be challenging when there is a limited amount of available data. Ensemble models are effective when the individual models being combined have the same number of parameters \cite{sagi2018ensemble}. Data augmentation introduces additional sources of variability, as there is no consensus on which transformations to apply, and the parameters of the transformation are often chosen randomly \cite{wen2020time}. The dearth of data makes ML and DL models overfit and increases model uncertainty. The results obtained in these scenarios may be a result of biased model performance, as the data do not capture the true underlying distribution. Consequently, analysis of the generality of the ML model across multiple datasets becomes extremely important to understand the true capability of the proposed model. With the growth of domain adaptation and transfer learning (TL) techniques, there seems to be an increase in the literature aiming to address the above-mentioned problems. TL aims to learn distributions over different domains and use the previously learned knowledge on the target task \cite{wan2021review}. Recent works on TL and its applications in different domains, including EEG processing, are mentioned in \cite{niu2020decade}. The use of TL has gained traction in the DL domain with the advancement of brain-computer interface (BCI) applications that utilize the knowledge learned from EEG data collected from different subjects across different sessions/tasks. Though fine-tuning pre-trained models for improved performance is one of the most widely used transfer learning techniques \cite{zhang2021adaptive}, there is no publicly available pre-trained model for EEG classification, unlike those present for image classification \cite{krizhevsky2012imagenet}. Our literature review identified four major works that involved TL between animals and humans, which are in line with the work showcased in this paper. In \cite{van2020transfer}, a robust deep neural network was proposed for electrocardiogram (ECG) beat classification. After the use of TL from the MIT-BIH dataset to the equine ECG (eECG) database, they showed an increase in performance metrics. In \cite{leroux2021chimpanzee}, the authors present an approach to leverage a feature extractor based on a DL network trained on human voice prints to provide an informative space over which chimpanzee voice prints can be recognized. Similarly, the primary focus in \cite{pandeya2018domestic} is the automatic classification of unseen domestic cat sounds using a pre-trained neural network. The authors in \cite{ntalampiras2018bird} proposed to transfer knowledge existing in music genre classification to identify bird species motivated by the existing acoustic similarities. As can be observed, all these works involve TL from humans to animal models, and most of these works involve pre-training DL models with human data and then using them to classify animal data. In contrast, in this paper, we propose to tackle the problem of the dearth of human biomedical data for training and increase the generalizability of the classification model by training the model on the Euclidean space-transformed animal (mice) data and classifying human data. 

DL has been extensively used in domains such as computer vision \cite{rawat2017deep,voulodimos2018deep} and speech recognition \cite{nassif2019speech} to yield improved performance compared to classical ML algorithms. However, its use in understanding phenomena in EEG is yet to be explored to its true potential. S. Gong et al. \cite{gong2021deep} have extensively reviewed the adoption of DL models in EEG domains. Thus far, a number of DL models such as Convolutional Neural Networks (CNN), Recurrent Neural Network (RNN), Autoencoders, and Generative Adversarial Network (GAN) have been used to model EEG data \cite{vo2022composing} and in various classification tasks such as motor imagery, sleep stage classification \cite{loh2020automated}, emotion recognition, and brain-computer interface \cite{lotte2018review}, with the most prevalent architecture being CNN followed by Deep Belief Networks (DBN) and hybrid architectures \cite{craik2019deep}. EEGNet encapsulates several well-known EEG feature extraction concepts \cite{lawhern2018eegnet}. Nonetheless, due to the low signal-to-noise ratio of the EEG signals, it sometimes becomes extremely difficult for the DL models to make any sense of the input signals. Cerasa et al. \cite{cerasa2022predicting} concluded in their study of comparison between traditional regression models and ML models for the same brain injury tasks that ML models did not provide significant improvement in the results although they helped in capturing better their non-linear relations. As a result, with the widespread reliance on DL models, one has to dwell deeper into the understanding of its advantages on EEG-related tasks for the extraction of meaningful results. Advances in DL architectures have renewed interest in multimodal prediction and classification of TBI \cite{lai2020detection,noor2020machine,thanjavur2021recurrent}.

In this work, our first aim is to explore different classical ML models such as decision trees (DT), random forest (RF), support vector machines (SVM), k-nearest neighbors (kNN), and extreme gradient boost (XGB) as well as EEGNet-based CNN model to detect mild TBI subjects. The robustness of the proposed models is tested by evaluating them on different datasets, including human and mouse data. We then aim to introduce  Transfer Euclidean Alignment (TEA) - a transfer alignment technique motivated by Correlation Alignment (CORAL) \cite{sun2016return} and Euclidean Alignment \cite{he2019transfer} to leverage the knowledge learned on mouse data to improve the classification accuracy on human data. The subsequent sections in this paper are organized as follows. In Section \ref{Methods}, we explain the mice and human datasets used in this study (Section \ref{Datasets}) and common EEG signal preprocessing steps involved in both rule-based ML and DL approaches (Section \ref{Preprocessing}). We then outlay the rule-based ML (Section \ref{ML}) and EEGNet-based DL (Section \ref{DL}) approach in detail. Finally, we discuss the results in Section \ref{Results} and draw our conclusion from this study in Section \ref{Conclusion}. 

\section{Methods}\label{Methods}
\subsection{Datasets}\label{Datasets}
\subsubsection{Mouse Data}

\underline{Dataset 1 ($\mathcal{D}_{1}$):} Mouse dataset 1 was acquired as previously published in \cite{lim2013dietary,modarres2017eeg}. Male C57BL/6J mice (Jackson Laboratory) at 10 weeks of age were housed in a laboratory environment with an ambient temperature of 23±1°C and a 12-hour light/12-hour dark cycle, which was automatically controlled. The animals were assigned to one of two groups: TBI and sham. Fluid Percussion Injury (FPI) along with EEG/EMG implantation in mice (n=12) was performed as described previously \cite{lim2013dietary}. Once the hub was FPI-induced, a 20-ms pulse of saline was delivered onto the dura with the pressure level between 1.4 and 2.1 atm \cite{carbonell1998adaptation,mcintosh1989traumatic}. The procedure was identical for sham mice, except for the introduction of the fluid pulse. EEG was recorded for five continuous days after the seven-day recovery period with a sampling frequency of 256 Hz. An experienced and blinded scorer performed sleep scoring, dividing the data into 4-second epochs of wakefulness (W), non-rapid eye movement (NREM), and rapid eye movement (REM) as previously described in \cite{modarres2017eeg}. 


\noindent
\underline{Dataset 2 ($\mathcal{D}_{2}$)}: Mouse dataset 2 was acquired as previously published in \cite{willie2012controlled}. Male C57BL6 mice (Jackson Laboratories, Bar Harbor, ME) at 4-6 months of age were used for the experiments. Once the surgical implantation of EEG and EMG electrodes for polysomnographic recordings were performed, left craniotomy and electromagnetically controlled cortical impact (CCI) was induced using a flat metal tip impounder driven at a velocity of 5 m/sec by an electromagnetic device or left craniotomy only (sham) was performed ipsilateral to the implanted probes following a 2-week recovery and rehabilitation period (n=5 mice per group) \cite{brody2007electromagnetic}. The electronically saved EEG/EMG records were initially scored using sleep scoring software (SleepSign; Kissei Comtec Co., Ltd., Nagano, Japan) into 4-sec epochs as W, REM, and NREM based on established criteria for rodent sleep \cite{renger2004sub}. The data were then manually over-scored by visual inspection and corrected when appropriate by a single investigator blinded to the intervention.

\subsubsection{Human Data}
Data from participants who underwent clinically indicated overnight polysomnography at a single site, VA Portland Health Care System (VAPORHCS) was accessed retrospectively under IRB approval (MIRB \#4108). A limited dataset of deidentified data from a subset of participants with a history of traumatic brain injury (TBI) and age-matched controls without TBI was institutionally transferred from VAPORHCS to UCIrvine under an approved Data Use Agreement. The source of data included age, sex, TBI status, and raw data from overnight polysomnography (PSG) using Polysmith (NihonKohden, Japan), with six scalp electrodes placed according to the 10-20 system of EEG placement at F3, F4, C3, C4, O1, and O2.
A certified polysomnographic technician manually scored PSG data and analyzed the sleep staging data in 30-second epochs based on standard clinical criteria. Each epoch was classified into one of the five sleep stages, namely awake (W), REM, and NREM stages N1, N2, and N3. A board-certified sleep physician additionally validated the staging as part of the clinical scoring process at VAPORHCS. 

\subsection{Preprocessing EEG signal}\label{Preprocessing}
Preprocessing is a crucial and necessary step in EEG analysis as the signal is inherently noisy \cite{keil2014committee} and contains various components that are not of interest in this work, such as eye movement, ECG components, and muscle artifacts \cite{robbins2020sensitive}. Eye movement and ECG components are removed using independent component analysis (ICA), which is a signal processing technique used to separate underlying independent components from observed multivariate non-gaussian data \cite{thatcher2009history}. In order to filter out bad epochs based on amplitude range, variance, and channel deviation, thresholds based on the Z-score are calculated rather than having an absolute threshold value. A Z-score of $\pm3$ is used as a threshold to identify contaminated data \cite{nolan2010faster}. Preprocessing EEG signals commonly involves filtering to remove noise. While filtering in the time domain causes a slight phase shift, frequency filtering does not \cite{oppenheim1999discrete}. Additionally, faster algorithms for calculating Fourier Transform (FT) and Inverse Fourier Transform (IFT) make frequency domain filtering more efficient compared to time domain filtering, which involves convolution steps \cite{widmann2015digital,yael2018filter}. Therefore, the filtering procedures used in this study involved three main steps: (1) converting the EEG signal from the time domain to the frequency domain using FT, (2) multiplying the frequency domain signal with the window of the required frequency range, and (3) reconstructing the filtered signal back to the time domain using IFT. The details of the procedure have been laid out in \cite{vishwanath2021detection}

\subsection{Rule-Based classical ML approach}\label{ML}
\subsubsection{Feature extraction}

Domain knowledge is essential in developing rule-based classical ML models. The model's performance significantly relies on the chosen features, so it is important to carefully select relevant QEEG features based on the task at hand \cite{rapp2015traumatic}. \cite{prichep2012classification,thatcher1989eeg,thatcher2001estimation} discuss various QEEG discriminant functions used for TBI detection.

To extract relevant features from EEG signals, spectral features such as the average and relative power, slow: fast power ratios, frequency amplitude asymmetry, and phase-amplitude coupling, connectivity features such as coherence, phase difference, and phase locking value, time domain features such as Hjorth parameter and non-linear features such as spectral entropy are calculated. The average and relative power in prominent EEG frequency bands - delta (0.5 – 4Hz), theta (4 – 8Hz), alpha (8 – 12Hz), sigma (12 - 16Hz), and beta (12 – 35Hz) are calculated as the area under the power spectral density (PSD) determined using Welch's periodogram and their ratios to the total power respectively. Additionally, power ratios corresponding to $theta:alpha$, and $alpha1:alpha2$ are also evaluated. To investigate the relative change in power distribution across different brain regions, frequency amplitude asymmetry is calculated as differences in absolute power between pairs of electrodes \cite{thatcher1989eeg}. For inter and intra-hemisphere comparisons, it is calculated as in Equation \ref{eq:2} and \ref{eq:3} respectively. Modulating the amplitude of high-frequency oscillations by the phase of the low-frequency component of the signal is a strong indicator of cross-frequency coupling characterized by phase-amplitude coupling (PAC) \cite{munia2019time}. For an observed EEG signal $x(t)$, phase $\phi_{x_l}(t)$ of the signal filtered at lower frequency range $f_l$ and amplitude envelope $A_{x_h}(t)$ of the signal filtered at higher frequency $f_h$ is calculated using Hilbert transform. Then the normalized mean amplitude is calculated for each frequency bin given by Equation \ref{eq:4} where $\left\langle \; \right\rangle$ denotes mean operation. Coherence provides a method for analyzing the spatial relationships between the EEG signals observed from different regions of the brain given by Equation \ref{eq: 5} where $S$ is the spectral-temporal density function of the signal. To overcome the shortcoming of coherence calculation, phase locking value (PLV) is calculated, which, unlike coherence, takes only the phase of the signals into consideration \cite{bruna2018phase} given by Equation \ref{eq:6} where $\phi_{1}$ and $\phi_{2}$ are the phase of the two EEG signals respectively. Hjorth parameters consist of three measures - activity, mobility, and complexity \cite{hjorth1970eeg}. Activity is calculated as the variance of the amplitude of the EEG signal (Equation \ref{eq:7}). Mobility is defined as the square root of the ratio between the variances of the first derivative and the amplitude of the EEG signal (Equation \ref{eq:8}), and complexity is a dimensionless quantity measured as the ratio between the mobility of the first derivative of the signal to the signal itself (Equation \ref{eq:9}). Non-linear features such as spectral entropy are calculated using standard formula \ref{eq:10} where $P(f)$ is normalized PSD and $f_{s}$ is the sampling frequency. All the measures calculated as input features for rule-based ML models are detailed in \cite{vishwanath2021detection}.

\begin{table}[t]
\begin{equation}
Relative\;Power=\frac{Power\;in\;frequency\;band}{Total\;Power}\label{eq:1}
\end{equation}

\begin{equation}
Freq\;Amplitude\;Asymmetry=\frac{Left\;-\;Right}{Left\;+\;Right}\label{eq:2}
\end{equation}

\begin{equation}
Freq\;Amplitude\;Asymmetry=\frac{Anterior\;-\;Posterior}{Anterior\;+\;Posterior}\label{eq:3}
\end{equation}

\begin{equation}
P(j)=\frac{\left\langle A_{x_{h}}\right\rangle_{\phi_{x_{l}}}(j)}{\sum_{k=1}^{N}\left\langle A_{x_{h}}\right\rangle_{\phi_{x_{l}}}(k)}\label{eq:4}
\end{equation}

\begin{equation}
{Coh}(f, t)=\frac{\left|\sum_{n} S_{1 n} \cdot S_{2 n}^{\prime}\right|^{2}}{\sum_{n}\left|S_{1,}\right|^{2} \cdot \sum_{n}\left|S_{2 n}\right|^{2}}\label{eq: 5}
\end{equation}

\begin{equation}
P L V(f, t)= \left| \frac{1}{N} \sum_{n} e^{i\left(\phi_{1 n}-\phi_{2 n}\right)}\right|\label{eq:6}
\end{equation}

\begin{equation}
{ Activity }(x(t))={var}(x(t)) \label{eq:7}
\end{equation}

\begin{equation}
{Mobility}(x(t))=\sqrt{\frac{{var}\left(\frac{d x(t)}{d t}\right)}{{var}(x(t))}} \label{eq:8}
\end{equation}

\begin{equation}
{ Complexity }(x(t))=\frac{ { Mobility }\left(\frac{d x(t)}{d t}\right)}{ { Mobility }(x(t))} \label{eq:9}
\end{equation}

\begin{equation}
H(x, s f)=-\sum_{f=0}^{f_{s} / 2} P(f) \log _{2}[P(f)]\label{eq:10}
\end{equation}
\end{table}

\subsubsection{Feature Normalization and Selection}
The calculated features then undergo a normalization procedure, which involves log transformation to overcome the skewness present in data, age regression to minimize the effect of age, and finally, Z-score standardization to bring all features to a similar scale of zero mean and unit standard deviation. Relative band power $R$ is transformed using $log(R/(1-R))$, magnitude squared coherence $C$ is transformed with $log(C/(1-C))$, amplitude asymmetry $X$ is transformed with $log((2+X)/(2-X))$ and spectral entropy $SpEn$ using --$log(1-SpEn)$ \cite{gasser1982transformations,john1987normative}. For age regression, the model assumes a linear relationship between the calculated features and $log_{10}$ of the subject's age expressed in years. The intercept and the coefficients obtained for the norming group (group of all control/normal subjects) are used to age regress the corresponding features of TBI subjects \cite{prichep2012classification} using the equation below.
\begin{equation}
y_{i}=x_{i}-\log 10( { SubjectAge }) \cdot m_{i}\label{eq:11}
\end{equation}
where $x_{i}$ and $y_{i}$ represent untransformed and transformed variables respectively and $m_{i}$ represents the age-regression parameter.

Selecting relevant features is a crucial step in developing an efficient ML model since it helps prevent overfitting and reduces the computational time and cost of model training. In this work, recursive feature elimination (RFE), which is a wrapper approach \cite{kohavi1997wrappers} to feature selection, is used. It iteratively evaluates the performance of an ML model on a subset of features based on which a decision is made to either add or remove features. RFE method with the random forest as the base estimator is used here.

\begin{figure*}
  \centering
  \includegraphics[width=\textwidth]{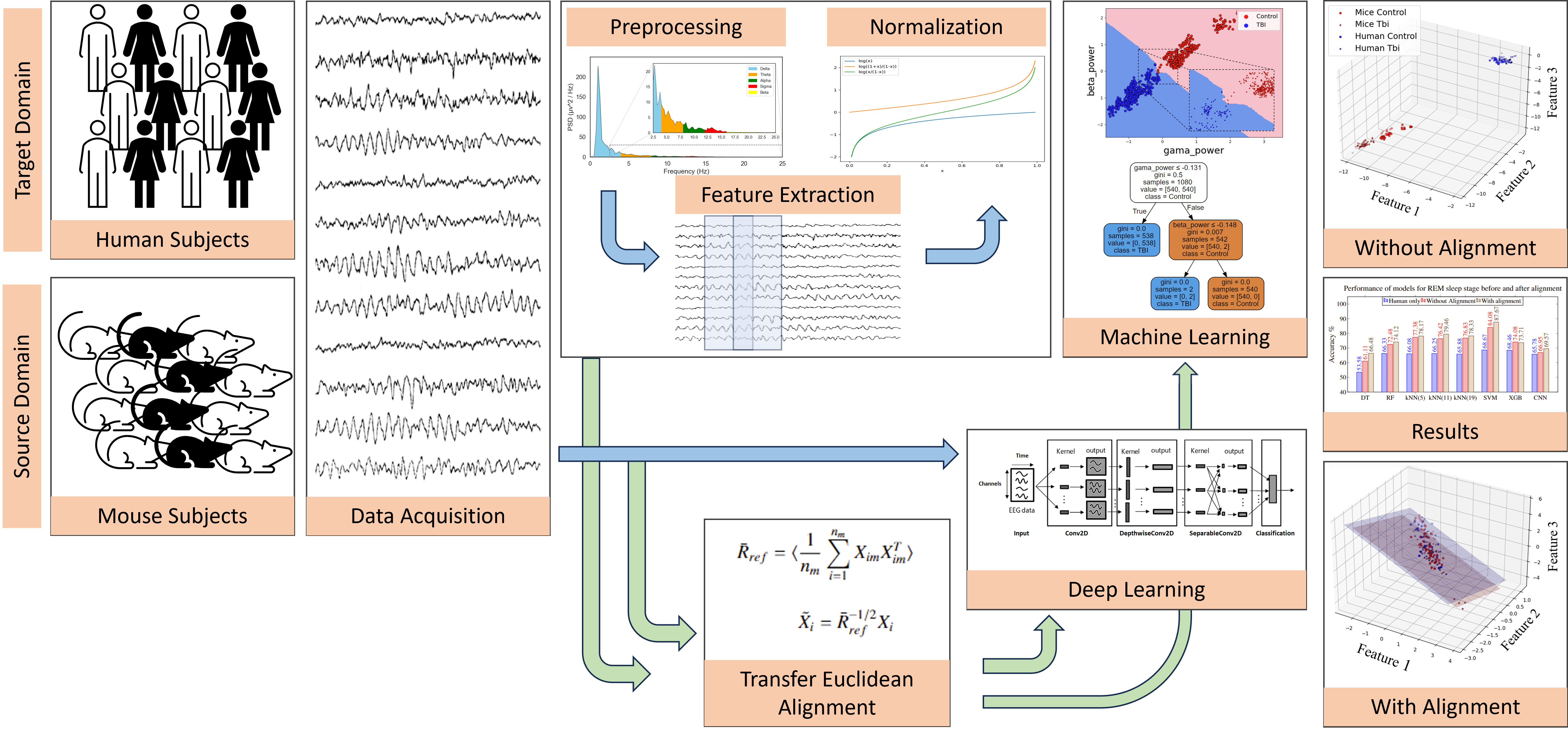}
  \caption{Overveiw of procedure. The EEG signals acquired from human and mouse subjects undergo binary classification (TBI/Control) either through rule-based ML models after feature extraction or through an EEGNet-based CNN model. The results obtained without the TEA pipeline (depicted in blue arrow) are then compared with the results obtained with the TEA pipeline (depicted in green arrow), along with an analysis of their feature space. }
  \label{Proc}
\end{figure*}

\subsubsection{Feature Alignment}\label{Feature Alignment}
As mentioned earlier, this work is motivated by unsupervised domain adaptation techniques - Correlation Alignment (CORAL) \cite{sun2016return} and Euclidean Alignment \cite{he2019transfer} that aims to reduce the difference between data distributions and align EEG trails in the Euclidean space enabling the use of machine learning algorithms that can be directly applied to the transformed trials. CORAL minimizes domain shifts between different datasets by aligning their second-order statistics. This is done by re-coloring whitened source features with the covariance of the target distribution. Unlike the work done in \cite{he2019transfer}, we aim to reduce the differences between different datasets mentioned in Subsection \ref{Datasets} and not between different subjects. As a result, we calculate the mean covariance of the entire dataset. In \cite{sun2016return}, the covariance of the dataset is considered; however, since our datasets consist of different subjects, we align the mean covariance matrix instead of the covariance matrix of the dataset. In \cite{sun2016return}, the authors choose to recolor the source dataset with the covariance of the target dataset; however, we follow the concept used in \cite{he2019transfer} to transform mean covariance matrices of all datasets into the identity matrix.

Consider dataset $\mathcal{D}_{1}$ containing a total of $m$ subjects and subject $m$ having $n_m$ trials. Mean covariance matrix $\bar{R}_{ref}$ for the entire dataset $\mathcal{D}_{1}$ is calculated using 
\begin{equation}
\bar{R}_{ref}=\langle \frac{1}{n_m} \sum_{i=1}^{n_m} X_{im} X_{im}^T \rangle\label{eq:12}
\end{equation}
where $X_{im}$ refers to the $ith$ trial of the $mth$ subject and $\langle \; \rangle$ denotes average across all $m$ subjects in the dataset. Giving importance to individual subject differences, $\bar{R}_{ref}$ is calculated as the average of covariances of individual subjects. Then, the alignment is performed using 
\begin{equation}
\tilde{X}_i=\bar{R}_{ref}^{-1 / 2} X_i\label{eq:13}
\end{equation}
This is carried out for all datasets, which reduces the mean covariance matrices of all datasets to the identity matrix as shown in Equation \ref{indentity}. Thus, making the distributions of the covariance matrices of different datasets similar. For classical rule-based ML algorithms, these operations are performed on the input feature space. 
\begin{equation}
\begin{split}
\langle \frac{1}{n_m} \sum_{i=1}^{n_{m}} \tilde{X}_{im} \tilde{X}_{im}^T \rangle=\langle \frac{1}{n_{m}} \sum_{i=1}^{n_{m}} \bar{R}_{ref}^{-1 / 2} X_i X_i^T \bar{R}_{ref}^{-1 / 2} \rangle
\\ =\bar{R}_{ref}^{-1 / 2} \langle \frac{1}{n_{m}} \sum_{i=1}^{n_{m}} X_i X_i^T \rangle \bar{R}_{ref}^{-1 / 2}=\bar{R}_{ref}^{-1 / 2} \bar{R}_{ref} \bar{R}_{ref}^{-1 / 2}=I
\label{indentity}
\end{split}
\end{equation}

\subsubsection{ML implementation}\label{ML implementation}
The calculated TEA features are fed into different rule-based ML models such as DT, RF, SVM, kNN, and XGB. The dataset is split in such a way that there are always 2 subjects in the test set - one from each class (control and TBI), and the data from the rest of the subjects are considered as a training set. This setup is known as independent validation and is repeated for different combinations of the train/test split. The model for kNN is run for 3 different values of "k" - 5, 11, and 19 to compare the significance of the number of neighbors in our classification task. SVM uses the radial basis function (RBF) kernel. All rule-based ML models are implemented using "sklearn" which is an ML module for Python. The overview of the classification procedure is shown in Fig. \ref{Proc}.

\begin{figure*}[t]
  \centering
  \includegraphics[width=\linewidth]{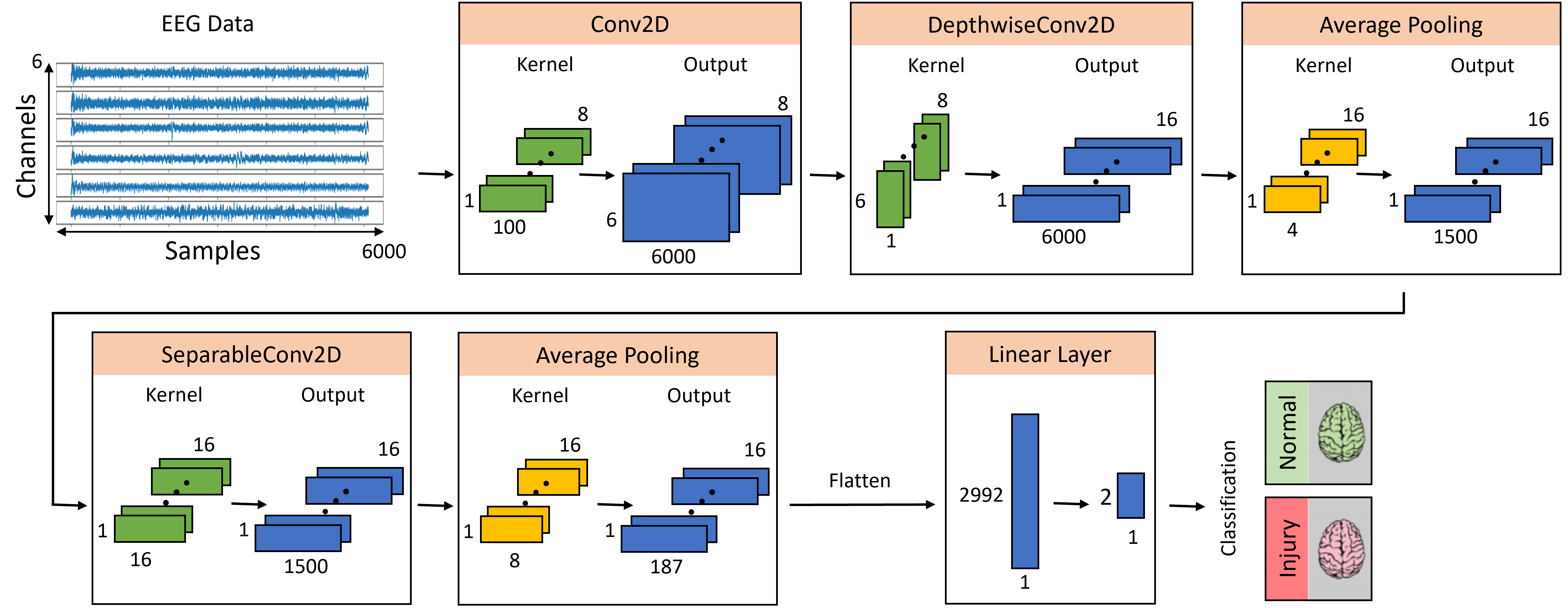}
  \caption{EEGNet-based CNN architecture.}
  \label{EEGNet}
\end{figure*}

\subsection{EEGNet-based DL Approach}\label{DL}
\subsubsection{Data Preparation and Alignment}
\begin{figure}[t]
  \centering
  \includegraphics[width=.9\linewidth]{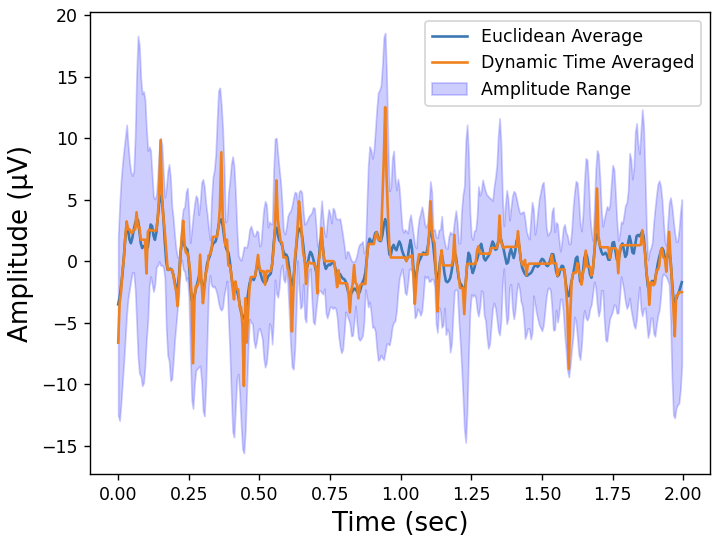}
  \caption{2-second window of Arithmetic averaged (blue) and  Dynamic Time Warping Averaged (orange; used in this work) EEG signal. Arithmetic averaging across 6 EEG channels will lead to loss of high-frequency information. The blue-shaded region depicts the range of EEG values calculated from the max and min of the 6 channels.}
  \label{mean}
\end{figure}

The preprocessed clean data obtained using the steps described in Section \ref{Preprocessing} are reshaped into tensors of the dimension 1 $\times$ 7168 and 6 $\times$ 6000 (number of channels $\times$ number of samples/epoch) for mice and humans respectively. Mice EEG typically features a single channel, in contrast to human EEG, which utilizes six channels as described in Section \ref{Datasets}. Further differentiating the two, mice EEG data are sampled at 256 Hz, whereas human EEG recordings are sampled at 200 Hz. Consequently, given a 30-second epoch length for human data and 28 seconds for mice, the input dimensions for the model are shaped as 1 $\times$ 7168 for mice and 6 $\times$ 6000 for humans. In the experiments involving TEA, the mouse data were resampled to 200 Hz, and the human data were averaged across channels using dynamic time warping averaging \cite{petitjean2014dynamic} to make the dimensions of the input data more consistent across species. Arithmetic averaging of time series data results in loss of information, especially high-frequency information. This is overcome by using dynamic time-warping averaging. Fig. \ref{mean} shows the arithmetic averaged (blue) as well as dynamic time warping averaged (orange) signals of a 2-second window of the 6 EEG channels (depicted as amplitude range) used in this study.  Then, the mean covariance matrices of both source (mice) and target (human) datasets were aligned together as discussed in Section \ref{Feature Alignment} so that they can be compared in a common Euclidean space.  First, we calculated the reference covariance matrix as the mean covariance matrix for each EEG dataset by averaging the covariance matrices of all subjects across all epochs. The covariance matrix of each dataset then underwent a transformation, resulting in their standardization to an identity matrix. Such standardization allows direct comparison and improves the accuracy of deep learning models when working with diverse datasets. 

\subsubsection{Model Architecture and Evaluation}

In this study, we employed a DL model derived from the EEGNet architecture \cite{lawhern2018eegnet}. EEGNet is a specialized compact CNN model designed explicitly for interpreting EEG data as its design incorporates specific techniques and methodologies tailored to EEG signals, such as feature extraction and optimal spatial filtering, consequently minimizing the number of parameters the model needs to learn. Here, we introduce a refined EEGNet architecture tailored to accommodate both human and mouse EEG data. While the layers of the architecture remain consistent, a crucial distinction emerges when we consider the input shape. Beyond these input variations explained previously, the overarching structure and processing components of the model remain uniform across both datasets.

In our efforts to adapt and optimize the EEGNet model to detect mTBI subjects, we undertook a series of modifications. Notably, after iterative testing, we adjusted the temporal convolution's length in the initial layer from 64 to 128, set the hyperparameter learning rate to 0.009, modified the number of epochs to 100, and tweaked the number of channels from 64 to 6 or 1 depending on the data input. Although we experimented with other changes such as the incorporation of a GlobalAveragePooling2D layer, fluctuating the dropout rate between 40\% and 60\%, and modifying the order of layers sequencing, these adjustments did not considerably improve the model performance. In addition, we added a callback function designed to adjust the learning rate according to the training loss dynamically. This involved the introduction of a 'patience' hyperparameter, dictating the epochs of stagnant loss values before halving the learning rate. Detailed architecture is shown in Fig. \ref{EEGNet}. The above-introduced model was trained using the independent validation method mentioned in Section \ref{ML implementation} to keep the evaluation method consistent between all explored ML and DL architectures. During the training process, the model's hyperparameters, including learning rate, batch size, kernel size, and number of epochs, were fine-tuned. These hyper-parameter optimizations were made to find a balance, preventing the model from either overfitting or underfitting the data. For evaluating CNN architecture, the average accuracy from 15 such test subject combinations is reported to ensure that the model is not biased due to the random selection of test subjects.  In the case of TEA, the model was first trained on Euclidean-aligned mouse data, after which they were retrained on the aligned human dataset. In other words, the model was pre-trained on the mouse data before fine-tuning with the human dataset. 

\begin{figure}[!h]
  \centering
  \includegraphics[width=\linewidth]{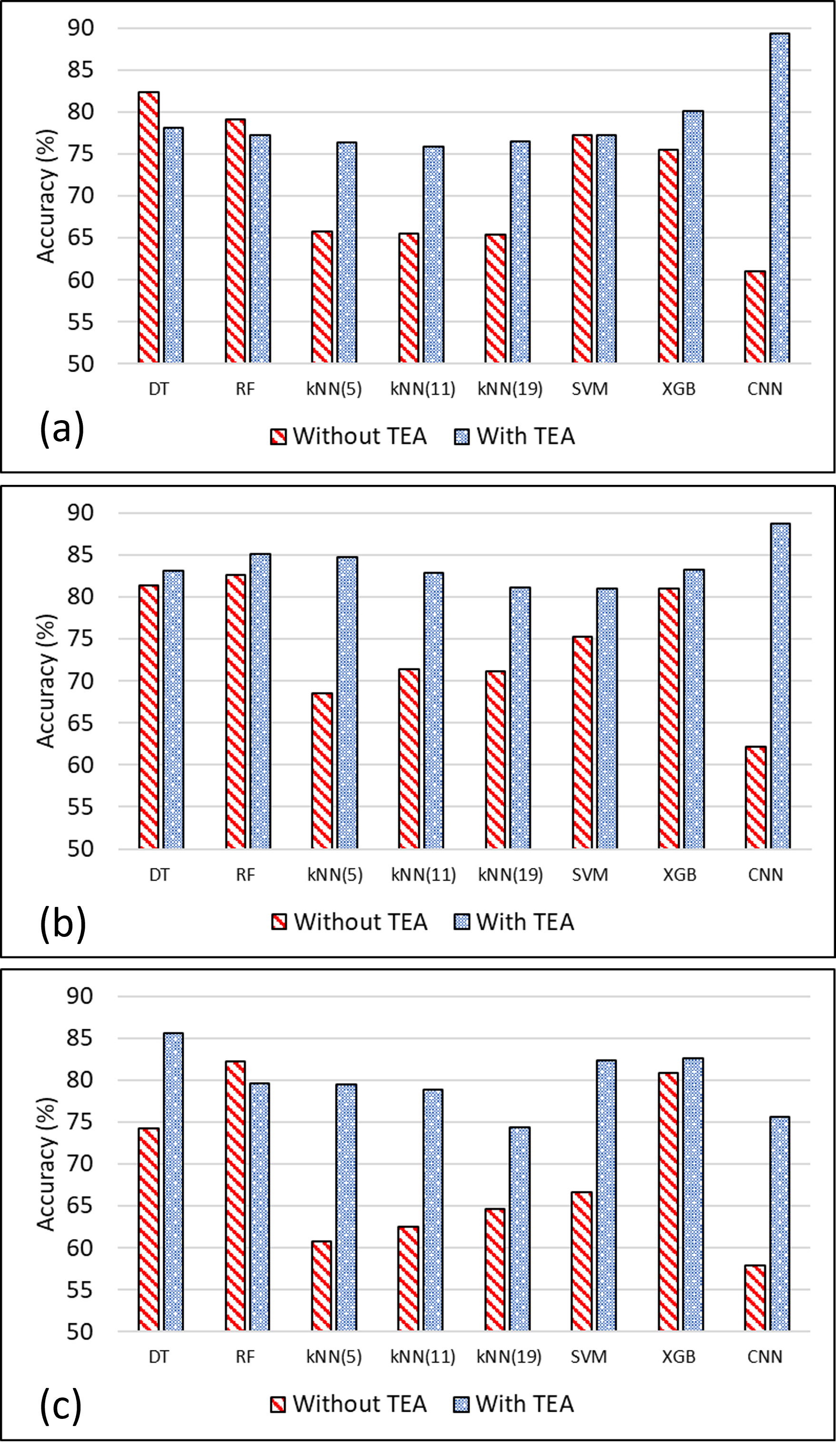}
  \caption{Performance of models for (a) W, (b) NREM, and (c) REM sleep stage before and after alignment when using only mouse datasets.}
  \label{graph1}
\end{figure}

\begin{figure}[!h]
  \centering
  \includegraphics[width=\linewidth]{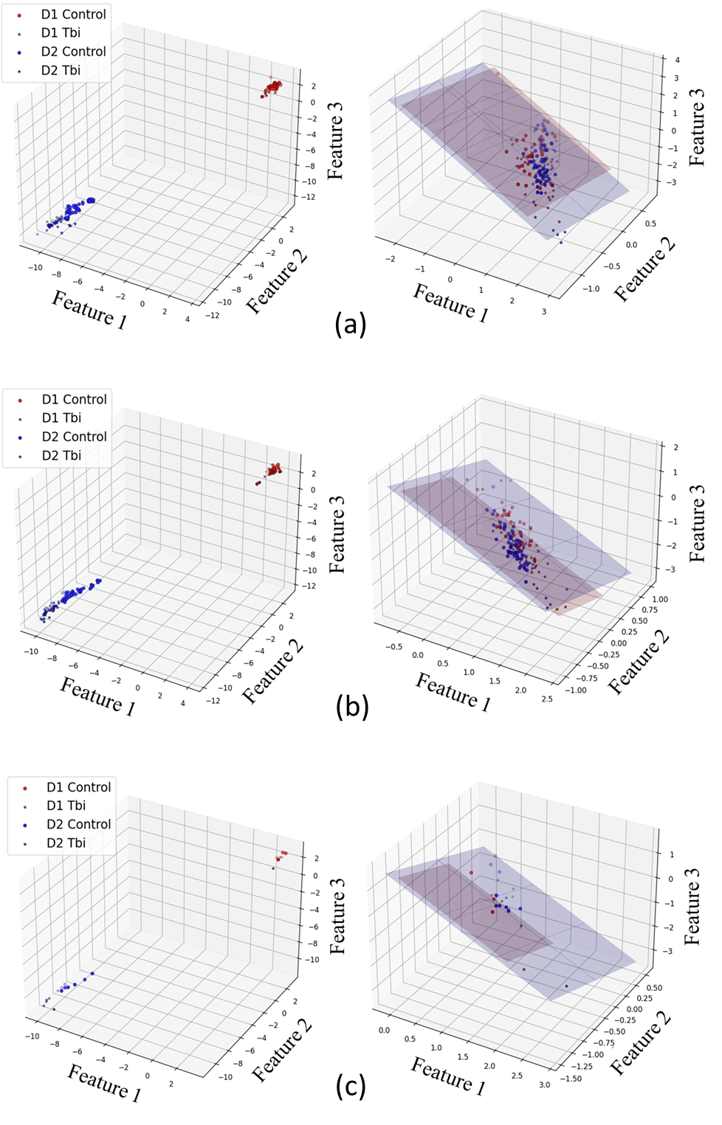}
  \caption{Feature space of mouse datasets before (left column) and after (right column) Transfer Euclidean Alignment, (a) for W, (b) NREM, and (c) REM sleep stages. Across all sleep stages, though dataset $\mathcal{D}_{1}$ (red) and dataset $\mathcal{D}_{2}$ (blue) share the same feature space, they do not follow the same distribution (shown in the left column) highlighting the covariate shifts present between the datasets. The right column depicts the alignment of the covariance matrices of the datasets after TEA, enabling models trained on $\mathcal{D}_{2}$ to perform well on $\mathcal{D}_{1}$.}
  \label{FS1}
\end{figure}

\begin{figure*}[!h]
  \centering
  \includegraphics[width=0.8\textwidth]{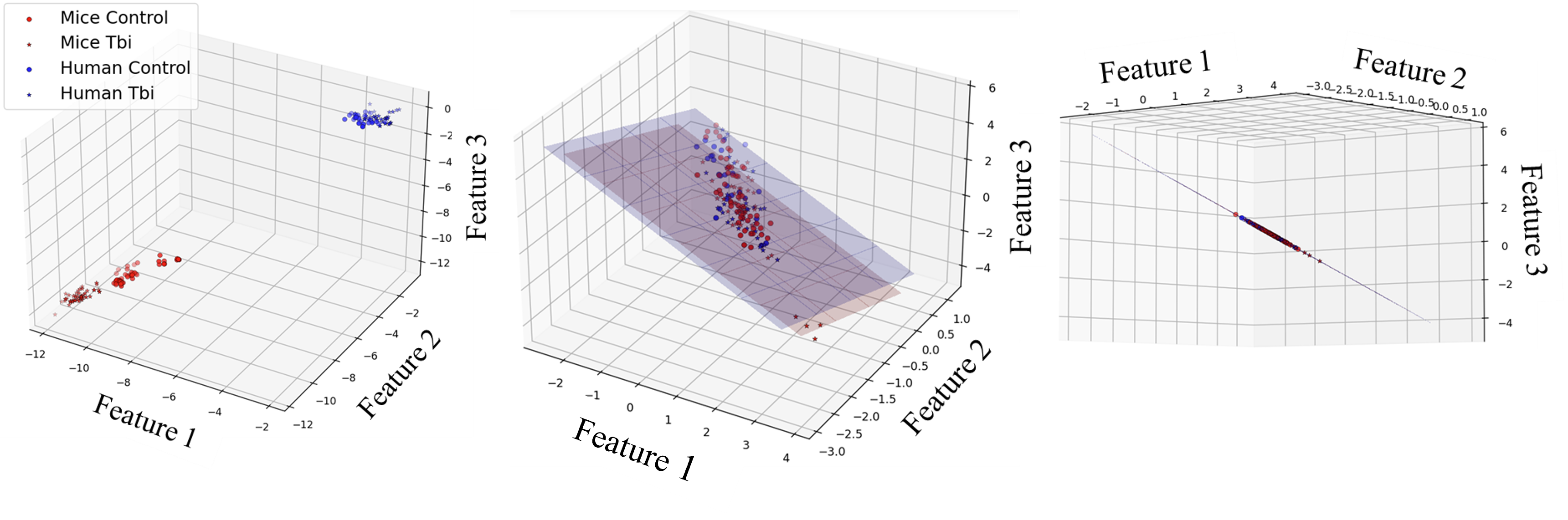}
  \caption{Feature space of mouse dataset (red) and human (blue) dataset before (left column) and after (center and right column) Transfer Euclidean Alignment for Wake sleep stages. The right column is a different view of the center column. Similar to Fig \ref{FS1}, the datasets do not follow the same distribution (shown in the left column), highlighting the covariate shifts present between the datasets. The center and right columns depict the alignment of the covariance matrices of the datasets after TEA, enabling models trained on mouse datasets to perform well on human datasets.}
  \label{FS2}
\end{figure*}

\begin{figure}[!h]
  \centering
  \includegraphics[width=\linewidth]{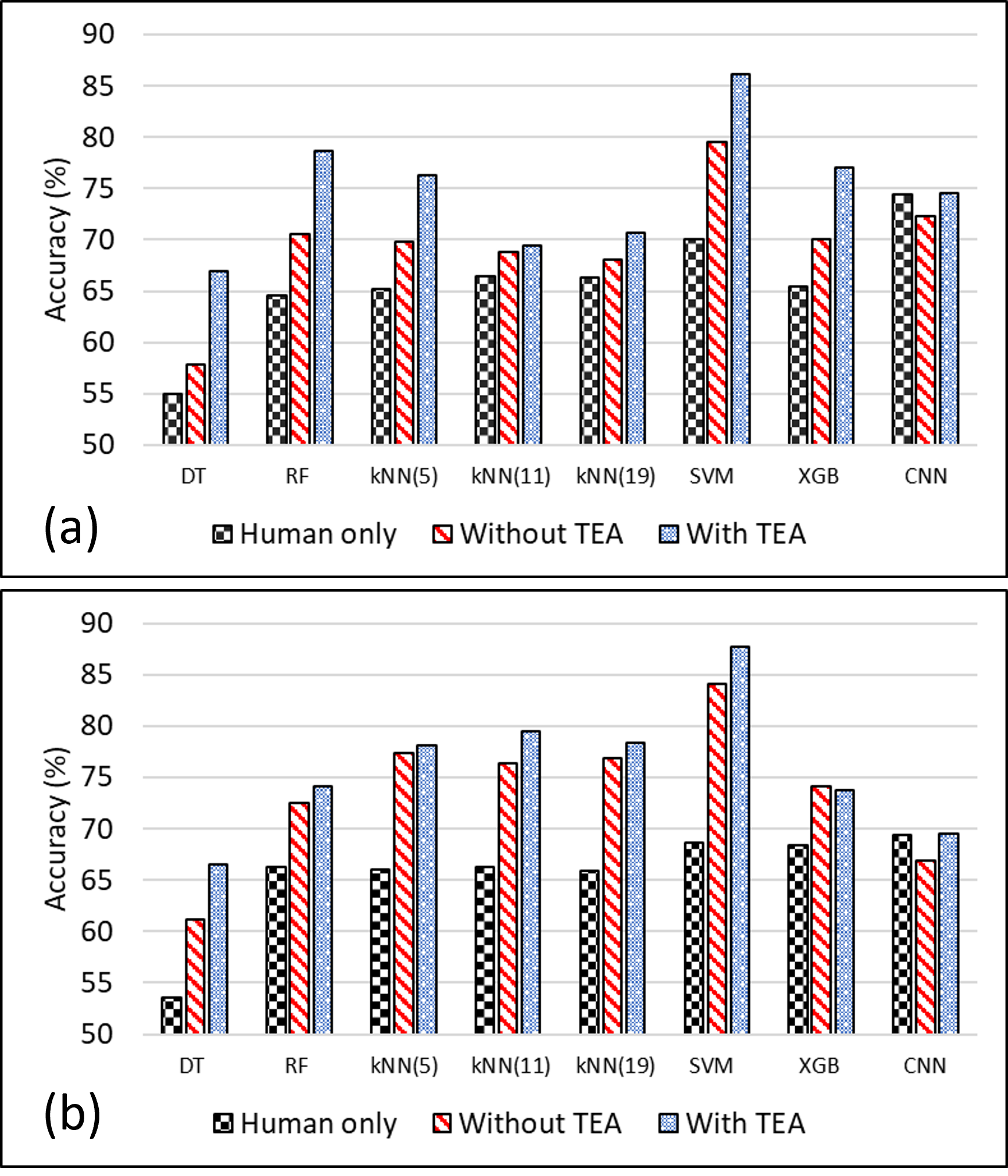}
  \caption{Performance of models for (a) W, and (b) REM sleep stage before and after alignment when using human and mouse datasets.}
  \label{graph2}
\end{figure}

\section{Results and Discussion}\label{Results}
There has been a significant amount of literature pointing toward studies that show convergent findings in mouse and human models of various neurological diseases. This becomes an important consideration in this work as its main objective is to benefit from the use of mice TBI EEG to train ML and DL models that can ultimately detect TBI accurately in human subjects. In a review study \cite{hafezparast2002mouse}, the authors answer some crucial questions on reasons that facilitate the modeling of neurological diseases in mice and why it is beneficial to use mice rather than other mammals. The authors in \cite{buzsaki2013scaling} and \cite{zhang2019cross} answer more fundamental questions of EEG conservation across species. Evolutionary preservation of brain rhythms, along with the stability of the constellation of the oscillation system in brains, is discussed in \cite{buzsaki2013scaling}.  Zhang et al. \cite{zhang2019cross} investigated the differences between eyes-closed and eyes-open conditions for humans and rats and concluded that the 1/f-like power spectrum scaling of resting state EEG activities was similar in both species. Focusing on TBI models in model organisms, Risling et al. \cite{risling2019modelling} discuss the advantages and disadvantages of animal research in TBI, specific to modeling TBI in rodent brains. The work done in this paper is primarily based on convergent findings in sleep-wake disturbances and EEG slow waves after TBI in mice and humans \cite{modarres2017eeg,sandsmark2017sleep}.

In the following subsections, we discuss the results obtained from 2 different experiments - Intra-species covariate shift reduction and Inter-species covariate shift reduction. Intra-species covariate shift reduction involves data from mice only. We intend to reduce the covariate shift between 2 different mouse datasets - $\mathcal{D}_{1}$ and $\mathcal{D}_{2}$ mentioned in Section \ref{Datasets} and thereby increase the performance of the models. In the experiment involving Inter-species covariate shift reduction, we use both the mice and human datasets. The mouse data are considered as the source domain dataset, and the human data are considered as the target domain. We leverage the mouse data to improve the classification of human data.

\subsection{Intraspecies Covariate Shift Reduction}

Mouse datasets $\mathcal{D}_{1}$ and $\mathcal{D}_{2}$ used in this experiment are considered to be the target and source datasets, respectively. The performance of the ML and DL models with and without an alignment is shown in Fig \ref{graph1}. The classification accuracy is shown on the y-axis, and the models compared in this study are shown on the x-axis. It is observed that the accuracy of almost all ML models across all sleep stages increases with alignment and more so for algorithms that are directly dependent on the distance between the instances/data points, such as kNN and SVM, with an average increase in accuracy of 23\% and 10\% respectively. DT and RF show relatively less improved performance after alignment in most conditions, which can be attributed to the nature of the ML model and its lesser reliance on distance metrics. The highest increase in performance is noticed in CNN, with an average increase of 40\%, which performs poorly before TEA when compared to ML models. Fig. \ref{FS1} shows the feature space of mouse datasets before (left column) and after (right column) TEA for different sleep stages. As it can be observed after TEA, the data points of the source dataset align to match the same Euclidean space as the target dataset, making it easier for ML models built on the source dataset to classify instances in the target domain. The shaded regions in the plots presented in the right column depict the planes fit to the source and target datasets, which coincide after using TEA.

\subsection{Interspecies Covariate Shift Reduction}
In this experiment, both mice and human data were used. To analyze the need for and the effect of TEA, we compare results across three different cases - (a) in the first case, only the human dataset was used to train and test the model, (b) in the second case the model was pre-trained using mouse dataset, and then fine-tuned using human training dataset, (c) the third case was similar to the second but involved the use of TEA datasets. As a result, the model was pre-trained using aligned mouse data and fine-tuned using aligned human training data. In all three cases, the test subjects used to evaluate the models were kept consistent so as to make a fair comparison. Fig \ref{graph2} shows the performance of the models across the wake and REM sleep stages. A consistent observation across all rule-based classical models is the increase in accuracy with the introduction of the pre-trained model using mouse data compared to using only human data. The use of TEA is also clearly seen with an increase in accuracy when aligned datasets are used across all models. The decrease in accuracy in the case of the DL model without alignment can be attributed to the difference in the data distribution and the relative ratio of the mice and human data available for the model to train on. Since the human data is PSG recording and the mouse data is a 24h recording the ratio of the amount of data available for both species is highly skewed, which might affect the performance of the model. It's also important to highlight that in CNN model, we utilize preprocessed EEG data as input, deviating from the manually crafted input features based on domain knowledge used in traditional rule-based ML models. However, this problem is overcome by the use of TEA, which shows an increase in performance when used. Fig. \ref{FS2} shows the feature space without and with TEA. Similar to Fig. \ref{FS1}, the datasets being used in this experiment align in the Euclidean space after the use of TEA, enabling the transfer of useful knowledge from the mouse dataset to the human classification task. A superior performance has been reported by the use of TEA compared to the previous works on the same datasets \cite{dhillon2021raspberry, vishwanath2020classification, vishwanath2020investigation, vishwanath2021investigation, vishwanath2021detection, vishwanath2022label}, validating the use of TEA to reduce covariate shifts between datasets. We suspect the use of TEA will require a much smaller human training dataset to achieve the same accuracy as using only human data, which will be explored in future work.

\section{Conclusion}\label{Conclusion}

In this study, we employed two distinct experiments to reduce covariate shifts between intra-species (mouse dataset) and inter-species (mouse and human datasets). Our objective was to evaluate the performance of ML and DL models in classifying TBI subjects from the control group. To the best of our knowledge, this is the first work demonstrating the use of transfer learning in Euclidean space that involves learning from a mouse model (that acts as the source data) to improve the performance of the target human dataset. We have effectively demonstrated the working of TEA in both classical rule-based ML settings as well as EEGNet-based DL models by showcasing significant enhancement in classification accuracy across all sleep stages. In conclusion, our research underscores the effectiveness of TEA as a transfer learning technique to bridge the gap between datasets with high subject-to-subject variability. Our findings suggest the promising potential of TEA in improving the generalizability and utility of machine learning models across diverse datasets, which has significant implications for real-world applications in the field of EEG anomaly detection and beyond.
\bibliographystyle{elsarticle-harv} 
\bibliography{main}

\begin{thebibliography}{74}
\expandafter\ifx\csname natexlab\endcsname\relax\def\natexlab#1{#1}\fi
\providecommand{\url}[1]{\texttt{#1}}
\providecommand{\href}[2]{#2}
\providecommand{\path}[1]{#1}
\providecommand{\DOIprefix}{doi:}
\providecommand{\ArXivprefix}{arXiv:}
\providecommand{\URLprefix}{URL: }
\providecommand{\Pubmedprefix}{pmid:}
\providecommand{\doi}[1]{\href{http://dx.doi.org/#1}{\path{#1}}}
\providecommand{\Pubmed}[1]{\href{pmid:#1}{\path{#1}}}
\providecommand{\bibinfo}[2]{#2}
\ifx\xfnm\relax \def\xfnm[#1]{\unskip,\space#1}\fi
\bibitem[{Acharya et~al.(2013)Acharya, Sree, Swapna, Martis and Suri}]{acharya2013automated}
\bibinfo{author}{Acharya, U.R.}, \bibinfo{author}{Sree, S.V.}, \bibinfo{author}{Swapna, G.}, \bibinfo{author}{Martis, R.J.}, \bibinfo{author}{Suri, J.S.}, \bibinfo{year}{2013}.
\newblock \bibinfo{title}{Automated eeg analysis of epilepsy: a review}.
\newblock \bibinfo{journal}{Knowledge-Based Systems} \bibinfo{volume}{45}, \bibinfo{pages}{147--165}.
\bibitem[{Brody et~al.(2007)Brody, Mac~Donald, Kessens, Yuede, Parsadanian, Spinner, Kim, Schwetye, Holtzman and Bayly}]{brody2007electromagnetic}
\bibinfo{author}{Brody, D.L.}, \bibinfo{author}{Mac~Donald, C.}, \bibinfo{author}{Kessens, C.C.}, \bibinfo{author}{Yuede, C.}, \bibinfo{author}{Parsadanian, M.}, \bibinfo{author}{Spinner, M.}, \bibinfo{author}{Kim, E.}, \bibinfo{author}{Schwetye, K.E.}, \bibinfo{author}{Holtzman, D.M.}, \bibinfo{author}{Bayly, P.V.}, \bibinfo{year}{2007}.
\newblock \bibinfo{title}{Electromagnetic controlled cortical impact device for precise, graded experimental traumatic brain injury}.
\newblock \bibinfo{journal}{Journal of neurotrauma} \bibinfo{volume}{24}, \bibinfo{pages}{657--673}.
\bibitem[{Bru{\~n}a et~al.(2018)Bru{\~n}a, Maest{\'u} and Pereda}]{bruna2018phase}
\bibinfo{author}{Bru{\~n}a, R.}, \bibinfo{author}{Maest{\'u}, F.}, \bibinfo{author}{Pereda, E.}, \bibinfo{year}{2018}.
\newblock \bibinfo{title}{Phase locking value revisited: teaching new tricks to an old dog}.
\newblock \bibinfo{journal}{Journal of neural engineering} \bibinfo{volume}{15}, \bibinfo{pages}{056011}.
\bibitem[{Buzs{\'a}ki et~al.(2013)Buzs{\'a}ki, Logothetis and Singer}]{buzsaki2013scaling}
\bibinfo{author}{Buzs{\'a}ki, G.}, \bibinfo{author}{Logothetis, N.}, \bibinfo{author}{Singer, W.}, \bibinfo{year}{2013}.
\newblock \bibinfo{title}{Scaling brain size, keeping timing: evolutionary preservation of brain rhythms}.
\newblock \bibinfo{journal}{Neuron} \bibinfo{volume}{80}, \bibinfo{pages}{751--764}.
\bibitem[{Carbonell et~al.(1998)Carbonell, Maris, McCALL and Grady}]{carbonell1998adaptation}
\bibinfo{author}{Carbonell, W.S.}, \bibinfo{author}{Maris, D.O.}, \bibinfo{author}{McCALL, T.}, \bibinfo{author}{Grady, M.S.}, \bibinfo{year}{1998}.
\newblock \bibinfo{title}{Adaptation of the fluid percussion injury model to the mouse}.
\newblock \bibinfo{journal}{Journal of neurotrauma} \bibinfo{volume}{15}, \bibinfo{pages}{217--229}.
\bibitem[{Cerasa et~al.(2022)Cerasa, Tartarisco, Bruschetta, Ciancarelli, Morone, Calabr{\`o}, Pioggia, Tonin and Iosa}]{cerasa2022predicting}
\bibinfo{author}{Cerasa, A.}, \bibinfo{author}{Tartarisco, G.}, \bibinfo{author}{Bruschetta, R.}, \bibinfo{author}{Ciancarelli, I.}, \bibinfo{author}{Morone, G.}, \bibinfo{author}{Calabr{\`o}, R.S.}, \bibinfo{author}{Pioggia, G.}, \bibinfo{author}{Tonin, P.}, \bibinfo{author}{Iosa, M.}, \bibinfo{year}{2022}.
\newblock \bibinfo{title}{Predicting outcome in patients with brain injury: differences between machine learning versus conventional statistics}.
\newblock \bibinfo{journal}{Biomedicines} \bibinfo{volume}{10}, \bibinfo{pages}{2267}.
\bibitem[{Craik et~al.(2019)Craik, He and Contreras-Vidal}]{craik2019deep}
\bibinfo{author}{Craik, A.}, \bibinfo{author}{He, Y.}, \bibinfo{author}{Contreras-Vidal, J.L.}, \bibinfo{year}{2019}.
\newblock \bibinfo{title}{Deep learning for electroencephalogram (eeg) classification tasks: a review}.
\newblock \bibinfo{journal}{Journal of neural engineering} \bibinfo{volume}{16}, \bibinfo{pages}{031001}.
\bibitem[{Dhillon et~al.(2021)Dhillon, Sutandi, Vishwanath, Lim, Cao and Si}]{dhillon2021raspberry}
\bibinfo{author}{Dhillon, N.S.}, \bibinfo{author}{Sutandi, A.}, \bibinfo{author}{Vishwanath, M.}, \bibinfo{author}{Lim, M.M.}, \bibinfo{author}{Cao, H.}, \bibinfo{author}{Si, D.}, \bibinfo{year}{2021}.
\newblock \bibinfo{title}{A raspberry pi-based traumatic brain injury detection system for single-channel electroencephalogram}.
\newblock \bibinfo{journal}{Sensors} \bibinfo{volume}{21}, \bibinfo{pages}{2779}.
\bibitem[{Gasser et~al.(1982)Gasser, B{\"a}cher and M{\"o}cks}]{gasser1982transformations}
\bibinfo{author}{Gasser, T.}, \bibinfo{author}{B{\"a}cher, P.}, \bibinfo{author}{M{\"o}cks, J.}, \bibinfo{year}{1982}.
\newblock \bibinfo{title}{Transformations towards the normal distribution of broad band spectral parameters of the eeg}.
\newblock \bibinfo{journal}{Electroencephalography and clinical neurophysiology} \bibinfo{volume}{53}, \bibinfo{pages}{119--124}.
\bibitem[{Gong et~al.(2021)Gong, Xing, Cichocki and Li}]{gong2021deep}
\bibinfo{author}{Gong, S.}, \bibinfo{author}{Xing, K.}, \bibinfo{author}{Cichocki, A.}, \bibinfo{author}{Li, J.}, \bibinfo{year}{2021}.
\newblock \bibinfo{title}{Deep learning in eeg: Advance of the last ten-year critical period}.
\newblock \bibinfo{journal}{IEEE Transactions on Cognitive and Developmental Systems} \bibinfo{volume}{14}, \bibinfo{pages}{348--365}.
\bibitem[{Hafezparast et~al.(2002)Hafezparast, Ahmad-Annuar, Wood, Tabrizi and Fisher}]{hafezparast2002mouse}
\bibinfo{author}{Hafezparast, M.}, \bibinfo{author}{Ahmad-Annuar, A.}, \bibinfo{author}{Wood, N.W.}, \bibinfo{author}{Tabrizi, S.J.}, \bibinfo{author}{Fisher, E.M.}, \bibinfo{year}{2002}.
\newblock \bibinfo{title}{Mouse models for neurological disease}.
\newblock \bibinfo{journal}{The Lancet Neurology} \bibinfo{volume}{1}, \bibinfo{pages}{215--224}.
\bibitem[{He and Wu(2019)}]{he2019transfer}
\bibinfo{author}{He, H.}, \bibinfo{author}{Wu, D.}, \bibinfo{year}{2019}.
\newblock \bibinfo{title}{Transfer learning for brain--computer interfaces: A euclidean space data alignment approach}.
\newblock \bibinfo{journal}{IEEE Transactions on Biomedical Engineering} \bibinfo{volume}{67}, \bibinfo{pages}{399--410}.
\bibitem[{Hjorth(1970)}]{hjorth1970eeg}
\bibinfo{author}{Hjorth, B.}, \bibinfo{year}{1970}.
\newblock \bibinfo{title}{Eeg analysis based on time domain properties}.
\newblock \bibinfo{journal}{Electroencephalography and clinical neurophysiology} \bibinfo{volume}{29}, \bibinfo{pages}{306--310}.
\bibitem[{Jeong(2004)}]{jeong2004eeg}
\bibinfo{author}{Jeong, J.}, \bibinfo{year}{2004}.
\newblock \bibinfo{title}{Eeg dynamics in patients with alzheimer's disease}.
\newblock \bibinfo{journal}{Clinical neurophysiology} \bibinfo{volume}{115}, \bibinfo{pages}{1490--1505}.
\bibitem[{John(1987)}]{john1987normative}
\bibinfo{author}{John, E.}, \bibinfo{year}{1987}.
\newblock \bibinfo{title}{Normative data bank and neurometrics. basic concepts, methods and results of norm constructions.}
\newblock \bibinfo{journal}{Methods od analysis of brain electrical and magnetic signals. EEG handbook} \bibinfo{volume}{1}, \bibinfo{pages}{449--498}.
\bibitem[{Kamrud et~al.(2021)Kamrud, Borghetti and Schubert~Kabban}]{kamrud2021effects}
\bibinfo{author}{Kamrud, A.}, \bibinfo{author}{Borghetti, B.}, \bibinfo{author}{Schubert~Kabban, C.}, \bibinfo{year}{2021}.
\newblock \bibinfo{title}{The effects of individual differences, non-stationarity, and the importance of data partitioning decisions for training and testing of eeg cross-participant models}.
\newblock \bibinfo{journal}{Sensors} \bibinfo{volume}{21}, \bibinfo{pages}{3225}.
\bibitem[{Keil et~al.(2014)Keil, Debener, Gratton, Jungh{\"o}fer, Kappenman, Luck, Luu, Miller and Yee}]{keil2014committee}
\bibinfo{author}{Keil, A.}, \bibinfo{author}{Debener, S.}, \bibinfo{author}{Gratton, G.}, \bibinfo{author}{Jungh{\"o}fer, M.}, \bibinfo{author}{Kappenman, E.S.}, \bibinfo{author}{Luck, S.J.}, \bibinfo{author}{Luu, P.}, \bibinfo{author}{Miller, G.A.}, \bibinfo{author}{Yee, C.M.}, \bibinfo{year}{2014}.
\newblock \bibinfo{title}{Committee report: publication guidelines and recommendations for studies using electroencephalography and magnetoencephalography}.
\newblock \bibinfo{journal}{Psychophysiology} \bibinfo{volume}{51}, \bibinfo{pages}{1--21}.
\bibitem[{Klimesch(1999)}]{klimesch1999eeg}
\bibinfo{author}{Klimesch, W.}, \bibinfo{year}{1999}.
\newblock \bibinfo{title}{Eeg alpha and theta oscillations reflect cognitive and memory performance: a review and analysis}.
\newblock \bibinfo{journal}{Brain research reviews} \bibinfo{volume}{29}, \bibinfo{pages}{169--195}.
\bibitem[{Kohavi and John(1997)}]{kohavi1997wrappers}
\bibinfo{author}{Kohavi, R.}, \bibinfo{author}{John, G.H.}, \bibinfo{year}{1997}.
\newblock \bibinfo{title}{Wrappers for feature subset selection}.
\newblock \bibinfo{journal}{Artificial intelligence} \bibinfo{volume}{97}, \bibinfo{pages}{273--324}.
\bibitem[{Krizhevsky et~al.(2012)Krizhevsky, Sutskever and Hinton}]{krizhevsky2012imagenet}
\bibinfo{author}{Krizhevsky, A.}, \bibinfo{author}{Sutskever, I.}, \bibinfo{author}{Hinton, G.E.}, \bibinfo{year}{2012}.
\newblock \bibinfo{title}{Imagenet classification with deep convolutional neural networks}.
\newblock \bibinfo{journal}{Advances in neural information processing systems} \bibinfo{volume}{25}.
\bibitem[{Lai et~al.(2020)Lai, Ibrahim, Abdullah, Azman, Abdullah et~al.}]{lai2020detection}
\bibinfo{author}{Lai, C.Q.}, \bibinfo{author}{Ibrahim, H.}, \bibinfo{author}{Abdullah, M.Z.}, \bibinfo{author}{Azman, A.}, \bibinfo{author}{Abdullah, J.M.}, et~al., \bibinfo{year}{2020}.
\newblock \bibinfo{title}{Detection of moderate traumatic brain injury from resting-state eye-closed electroencephalography}.
\newblock \bibinfo{journal}{Computational intelligence and neuroscience} \bibinfo{volume}{2020}.
\bibitem[{Lawhern et~al.(2018)Lawhern, Solon, Waytowich, Gordon, Hung and Lance}]{lawhern2018eegnet}
\bibinfo{author}{Lawhern, V.J.}, \bibinfo{author}{Solon, A.J.}, \bibinfo{author}{Waytowich, N.R.}, \bibinfo{author}{Gordon, S.M.}, \bibinfo{author}{Hung, C.P.}, \bibinfo{author}{Lance, B.J.}, \bibinfo{year}{2018}.
\newblock \bibinfo{title}{Eegnet: a compact convolutional neural network for eeg-based brain--computer interfaces}.
\newblock \bibinfo{journal}{Journal of neural engineering} \bibinfo{volume}{15}, \bibinfo{pages}{056013}.
\bibitem[{Leroux et~al.(2021)Leroux, Al-Khudhairy, Perony and Townsend}]{leroux2021chimpanzee}
\bibinfo{author}{Leroux, M.}, \bibinfo{author}{Al-Khudhairy, O.G.}, \bibinfo{author}{Perony, N.}, \bibinfo{author}{Townsend, S.W.}, \bibinfo{year}{2021}.
\newblock \bibinfo{title}{Chimpanzee voice prints? insights from transfer learning experiments from human voices}.
\newblock \bibinfo{journal}{arXiv preprint arXiv:2112.08165} .
\bibitem[{Lim et~al.(2013)Lim, Elkind, Xiong, Galante, Zhu, Zhang, Lian, Rodin, Kuzma, Pack et~al.}]{lim2013dietary}
\bibinfo{author}{Lim, M.M.}, \bibinfo{author}{Elkind, J.}, \bibinfo{author}{Xiong, G.}, \bibinfo{author}{Galante, R.}, \bibinfo{author}{Zhu, J.}, \bibinfo{author}{Zhang, L.}, \bibinfo{author}{Lian, J.}, \bibinfo{author}{Rodin, J.}, \bibinfo{author}{Kuzma, N.N.}, \bibinfo{author}{Pack, A.I.}, et~al., \bibinfo{year}{2013}.
\newblock \bibinfo{title}{Dietary therapy mitigates persistent wake deficits caused by mild traumatic brain injury}.
\newblock \bibinfo{journal}{Science translational medicine} \bibinfo{volume}{5}, \bibinfo{pages}{215ra173--215ra173}.
\bibitem[{Liu et~al.(2013)Liu, Chiang and Chu}]{liu2013recognizing}
\bibinfo{author}{Liu, N.H.}, \bibinfo{author}{Chiang, C.Y.}, \bibinfo{author}{Chu, H.C.}, \bibinfo{year}{2013}.
\newblock \bibinfo{title}{Recognizing the degree of human attention using eeg signals from mobile sensors}.
\newblock \bibinfo{journal}{sensors} \bibinfo{volume}{13}, \bibinfo{pages}{10273--10286}.
\bibitem[{Loh et~al.(2020)Loh, Ooi, Vicnesh, Oh, Faust, Gertych and Acharya}]{loh2020automated}
\bibinfo{author}{Loh, H.W.}, \bibinfo{author}{Ooi, C.P.}, \bibinfo{author}{Vicnesh, J.}, \bibinfo{author}{Oh, S.L.}, \bibinfo{author}{Faust, O.}, \bibinfo{author}{Gertych, A.}, \bibinfo{author}{Acharya, U.R.}, \bibinfo{year}{2020}.
\newblock \bibinfo{title}{Automated detection of sleep stages using deep learning techniques: A systematic review of the last decade (2010--2020)}.
\newblock \bibinfo{journal}{Applied Sciences} \bibinfo{volume}{10}, \bibinfo{pages}{8963}.
\bibitem[{Lotte et~al.(2018)Lotte, Bougrain, Cichocki, Clerc, Congedo, Rakotomamonjy and Yger}]{lotte2018review}
\bibinfo{author}{Lotte, F.}, \bibinfo{author}{Bougrain, L.}, \bibinfo{author}{Cichocki, A.}, \bibinfo{author}{Clerc, M.}, \bibinfo{author}{Congedo, M.}, \bibinfo{author}{Rakotomamonjy, A.}, \bibinfo{author}{Yger, F.}, \bibinfo{year}{2018}.
\newblock \bibinfo{title}{A review of classification algorithms for eeg-based brain--computer interfaces: a 10 year update}.
\newblock \bibinfo{journal}{Journal of neural engineering} \bibinfo{volume}{15}, \bibinfo{pages}{031005}.
\bibitem[{McIntosh et~al.(1989)McIntosh, Vink, Noble, Yamakami, Fernyak, Soares and Faden}]{mcintosh1989traumatic}
\bibinfo{author}{McIntosh, T.}, \bibinfo{author}{Vink, R.}, \bibinfo{author}{Noble, L.}, \bibinfo{author}{Yamakami, I.}, \bibinfo{author}{Fernyak, S.}, \bibinfo{author}{Soares, H.}, \bibinfo{author}{Faden, A.}, \bibinfo{year}{1989}.
\newblock \bibinfo{title}{Traumatic brain injury in the rat: characterization of a lateral fluid-percussion model}.
\newblock \bibinfo{journal}{Neuroscience} \bibinfo{volume}{28}, \bibinfo{pages}{233--244}.
\bibitem[{Modarres et~al.(2017)Modarres, Kuzma, Kretzmer, Pack and Lim}]{modarres2017eeg}
\bibinfo{author}{Modarres, M.H.}, \bibinfo{author}{Kuzma, N.N.}, \bibinfo{author}{Kretzmer, T.}, \bibinfo{author}{Pack, A.I.}, \bibinfo{author}{Lim, M.M.}, \bibinfo{year}{2017}.
\newblock \bibinfo{title}{Eeg slow waves in traumatic brain injury: Convergent findings in mouse and man}.
\newblock \bibinfo{journal}{Neurobiology of sleep and circadian rhythms} \bibinfo{volume}{2}, \bibinfo{pages}{59--70}.
\bibitem[{Munia and Aviyente(2019)}]{munia2019time}
\bibinfo{author}{Munia, T.T.}, \bibinfo{author}{Aviyente, S.}, \bibinfo{year}{2019}.
\newblock \bibinfo{title}{Time-frequency based phase-amplitude coupling measure for neuronal oscillations}.
\newblock \bibinfo{journal}{Scientific reports} \bibinfo{volume}{9}, \bibinfo{pages}{1--15}.
\bibitem[{Nassif et~al.(2019)Nassif, Shahin, Attili, Azzeh and Shaalan}]{nassif2019speech}
\bibinfo{author}{Nassif, A.B.}, \bibinfo{author}{Shahin, I.}, \bibinfo{author}{Attili, I.}, \bibinfo{author}{Azzeh, M.}, \bibinfo{author}{Shaalan, K.}, \bibinfo{year}{2019}.
\newblock \bibinfo{title}{Speech recognition using deep neural networks: A systematic review}.
\newblock \bibinfo{journal}{IEEE access} \bibinfo{volume}{7}, \bibinfo{pages}{19143--19165}.
\bibitem[{Niu et~al.(2020)Niu, Liu, Wang and Song}]{niu2020decade}
\bibinfo{author}{Niu, S.}, \bibinfo{author}{Liu, Y.}, \bibinfo{author}{Wang, J.}, \bibinfo{author}{Song, H.}, \bibinfo{year}{2020}.
\newblock \bibinfo{title}{A decade survey of transfer learning (2010--2020)}.
\newblock \bibinfo{journal}{IEEE Transactions on Artificial Intelligence} \bibinfo{volume}{1}, \bibinfo{pages}{151--166}.
\bibitem[{Noachtar and R{\'e}mi(2009)}]{noachtar2009role}
\bibinfo{author}{Noachtar, S.}, \bibinfo{author}{R{\'e}mi, J.}, \bibinfo{year}{2009}.
\newblock \bibinfo{title}{The role of eeg in epilepsy: a critical review}.
\newblock \bibinfo{journal}{Epilepsy \& Behavior} \bibinfo{volume}{15}, \bibinfo{pages}{22--33}.
\bibitem[{Nolan et~al.(2010)Nolan, Whelan and Reilly}]{nolan2010faster}
\bibinfo{author}{Nolan, H.}, \bibinfo{author}{Whelan, R.}, \bibinfo{author}{Reilly, R.B.}, \bibinfo{year}{2010}.
\newblock \bibinfo{title}{Faster: fully automated statistical thresholding for eeg artifact rejection}.
\newblock \bibinfo{journal}{Journal of neuroscience methods} \bibinfo{volume}{192}, \bibinfo{pages}{152--162}.
\bibitem[{Noor and Ibrahim(2020)}]{noor2020machine}
\bibinfo{author}{Noor, N.S.E.M.}, \bibinfo{author}{Ibrahim, H.}, \bibinfo{year}{2020}.
\newblock \bibinfo{title}{Machine learning algorithms and quantitative electroencephalography predictors for outcome prediction in traumatic brain injury: A systematic review}.
\newblock \bibinfo{journal}{IEEE Access} \bibinfo{volume}{8}, \bibinfo{pages}{102075--102092}.
\bibitem[{Ntalampiras(2018)}]{ntalampiras2018bird}
\bibinfo{author}{Ntalampiras, S.}, \bibinfo{year}{2018}.
\newblock \bibinfo{title}{Bird species identification via transfer learning from music genres}.
\newblock \bibinfo{journal}{Ecological informatics} \bibinfo{volume}{44}, \bibinfo{pages}{76--81}.
\bibitem[{Nunez et~al.(2006)Nunez, Srinivasan et~al.}]{nunez2006electric}
\bibinfo{author}{Nunez, P.L.}, \bibinfo{author}{Srinivasan, R.}, et~al., \bibinfo{year}{2006}.
\newblock \bibinfo{title}{Electric fields of the brain: the neurophysics of EEG}.
\newblock \bibinfo{publisher}{Oxford University Press, USA}.
\bibitem[{Oppenheim(1999)}]{oppenheim1999discrete}
\bibinfo{author}{Oppenheim, A.V.}, \bibinfo{year}{1999}.
\newblock \bibinfo{title}{Discrete-time signal processing}.
\newblock \bibinfo{publisher}{Pearson Education India}.
\bibitem[{Pan and Yang(2010)}]{pan2010survey}
\bibinfo{author}{Pan, S.J.}, \bibinfo{author}{Yang, Q.}, \bibinfo{year}{2010}.
\newblock \bibinfo{title}{A survey on transfer learning}.
\newblock \bibinfo{journal}{IEEE Transactions on knowledge and data engineering} \bibinfo{volume}{22}, \bibinfo{pages}{1345--1359}.
\bibitem[{Pandeya and Lee(2018)}]{pandeya2018domestic}
\bibinfo{author}{Pandeya, Y.R.}, \bibinfo{author}{Lee, J.}, \bibinfo{year}{2018}.
\newblock \bibinfo{title}{Domestic cat sound classification using transfer learning}.
\newblock \bibinfo{journal}{International Journal of Fuzzy Logic and Intelligent Systems} \bibinfo{volume}{18}, \bibinfo{pages}{154--160}.
\bibitem[{Petitjean et~al.(2014)Petitjean, Forestier, Webb, Nicholson, Chen and Keogh}]{petitjean2014dynamic}
\bibinfo{author}{Petitjean, F.}, \bibinfo{author}{Forestier, G.}, \bibinfo{author}{Webb, G.I.}, \bibinfo{author}{Nicholson, A.E.}, \bibinfo{author}{Chen, Y.}, \bibinfo{author}{Keogh, E.}, \bibinfo{year}{2014}.
\newblock \bibinfo{title}{Dynamic time warping averaging of time series allows faster and more accurate classification}, in: \bibinfo{booktitle}{2014 IEEE international conference on data mining}, \bibinfo{organization}{IEEE}. pp. \bibinfo{pages}{470--479}.
\bibitem[{Prichep(2005)}]{prichep2005use}
\bibinfo{author}{Prichep, L.S.}, \bibinfo{year}{2005}.
\newblock \bibinfo{title}{Use of normative databases and statistical methods in demonstrating clinical utility of qeeg: importance and cautions}.
\newblock \bibinfo{journal}{Clinical EEG and neuroscience} \bibinfo{volume}{36}, \bibinfo{pages}{82--87}.
\bibitem[{Prichep et~al.(2012)Prichep, Jacquin, Filipenko, Dastidar, Zabele, Vodencarevic and Rothman}]{prichep2012classification}
\bibinfo{author}{Prichep, L.S.}, \bibinfo{author}{Jacquin, A.}, \bibinfo{author}{Filipenko, J.}, \bibinfo{author}{Dastidar, S.G.}, \bibinfo{author}{Zabele, S.}, \bibinfo{author}{Vodencarevic, A.}, \bibinfo{author}{Rothman, N.S.}, \bibinfo{year}{2012}.
\newblock \bibinfo{title}{Classification of traumatic brain injury severity using informed data reduction in a series of binary classifier algorithms}.
\newblock \bibinfo{journal}{IEEE transactions on neural systems and rehabilitation engineering} \bibinfo{volume}{20}, \bibinfo{pages}{806--822}.
\bibitem[{Rapp et~al.(2015)Rapp, Keyser, Albano, Hernandez, Gibson, Zambon, Hairston, Hughes, Krystal and Nichols}]{rapp2015traumatic}
\bibinfo{author}{Rapp, P.E.}, \bibinfo{author}{Keyser, D.O.}, \bibinfo{author}{Albano, A.}, \bibinfo{author}{Hernandez, R.}, \bibinfo{author}{Gibson, D.B.}, \bibinfo{author}{Zambon, R.A.}, \bibinfo{author}{Hairston, W.D.}, \bibinfo{author}{Hughes, J.D.}, \bibinfo{author}{Krystal, A.}, \bibinfo{author}{Nichols, A.S.}, \bibinfo{year}{2015}.
\newblock \bibinfo{title}{Traumatic brain injury detection using electrophysiological methods}.
\newblock \bibinfo{journal}{Frontiers in human neuroscience} \bibinfo{volume}{9}, \bibinfo{pages}{11}.
\bibitem[{Rawat and Wang(2017)}]{rawat2017deep}
\bibinfo{author}{Rawat, W.}, \bibinfo{author}{Wang, Z.}, \bibinfo{year}{2017}.
\newblock \bibinfo{title}{Deep convolutional neural networks for image classification: A comprehensive review}.
\newblock \bibinfo{journal}{Neural computation} \bibinfo{volume}{29}, \bibinfo{pages}{2352--2449}.
\bibitem[{Renger et~al.(2004)Renger, Dunn, Motzel, Johnson and Koblan}]{renger2004sub}
\bibinfo{author}{Renger, J.J.}, \bibinfo{author}{Dunn, S.L.}, \bibinfo{author}{Motzel, S.L.}, \bibinfo{author}{Johnson, C.}, \bibinfo{author}{Koblan, K.S.}, \bibinfo{year}{2004}.
\newblock \bibinfo{title}{Sub-chronic administration of zolpidem affects modifications to rat sleep architecture}.
\newblock \bibinfo{journal}{Brain research} \bibinfo{volume}{1010}, \bibinfo{pages}{45--54}.
\bibitem[{Risling et~al.(2019)Risling, Smith, Stein, Thelin, Zanier, Ankarcrona and Nilsson}]{risling2019modelling}
\bibinfo{author}{Risling, M.}, \bibinfo{author}{Smith, D.}, \bibinfo{author}{Stein, T.D.}, \bibinfo{author}{Thelin, E.P.}, \bibinfo{author}{Zanier, E.R.}, \bibinfo{author}{Ankarcrona, M.}, \bibinfo{author}{Nilsson, P.}, \bibinfo{year}{2019}.
\newblock \bibinfo{title}{Modelling human pathology of traumatic brain injury in animal models}.
\newblock \bibinfo{journal}{Journal of internal medicine} \bibinfo{volume}{285}, \bibinfo{pages}{594--607}.
\bibitem[{Robbins et~al.(2020)Robbins, Touryan, Mullen, Kothe and Bigdely-Shamlo}]{robbins2020sensitive}
\bibinfo{author}{Robbins, K.A.}, \bibinfo{author}{Touryan, J.}, \bibinfo{author}{Mullen, T.}, \bibinfo{author}{Kothe, C.}, \bibinfo{author}{Bigdely-Shamlo, N.}, \bibinfo{year}{2020}.
\newblock \bibinfo{title}{How sensitive are eeg results to preprocessing methods: a benchmarking study}.
\newblock \bibinfo{journal}{IEEE Transactions on Neural Systems and Rehabilitation Engineering} \bibinfo{volume}{28}, \bibinfo{pages}{1081--1090}.
\bibitem[{Sagi and Rokach(2018)}]{sagi2018ensemble}
\bibinfo{author}{Sagi, O.}, \bibinfo{author}{Rokach, L.}, \bibinfo{year}{2018}.
\newblock \bibinfo{title}{Ensemble learning: A survey}.
\newblock \bibinfo{journal}{Wiley Interdisciplinary Reviews: Data Mining and Knowledge Discovery} \bibinfo{volume}{8}, \bibinfo{pages}{e1249}.
\bibitem[{Sandsmark et~al.(2017)Sandsmark, Elliott and Lim}]{sandsmark2017sleep}
\bibinfo{author}{Sandsmark, D.K.}, \bibinfo{author}{Elliott, J.E.}, \bibinfo{author}{Lim, M.M.}, \bibinfo{year}{2017}.
\newblock \bibinfo{title}{Sleep-wake disturbances after traumatic brain injury: synthesis of human and animal studies}.
\newblock \bibinfo{journal}{Sleep} \bibinfo{volume}{40}, \bibinfo{pages}{zsx044}.
\bibitem[{Sun et~al.(2016)Sun, Feng and Saenko}]{sun2016return}
\bibinfo{author}{Sun, B.}, \bibinfo{author}{Feng, J.}, \bibinfo{author}{Saenko, K.}, \bibinfo{year}{2016}.
\newblock \bibinfo{title}{Return of frustratingly easy domain adaptation}, in: \bibinfo{booktitle}{Proceedings of the AAAI Conference on Artificial Intelligence}.
\bibitem[{Thanjavur et~al.(2021)Thanjavur, Babul, Foran, Bielecki, Gilchrist, Hristopulos, Brucar and Virji-Babul}]{thanjavur2021recurrent}
\bibinfo{author}{Thanjavur, K.}, \bibinfo{author}{Babul, A.}, \bibinfo{author}{Foran, B.}, \bibinfo{author}{Bielecki, M.}, \bibinfo{author}{Gilchrist, A.}, \bibinfo{author}{Hristopulos, D.T.}, \bibinfo{author}{Brucar, L.R.}, \bibinfo{author}{Virji-Babul, N.}, \bibinfo{year}{2021}.
\newblock \bibinfo{title}{Recurrent neural network-based acute concussion classifier using raw resting state eeg data}.
\newblock \bibinfo{journal}{Scientific reports} \bibinfo{volume}{11}, \bibinfo{pages}{12353}.
\bibitem[{Thatcher et~al.(2001)Thatcher, Biver, Gomez, North, Curtin, Walker and Salazar}]{thatcher2001estimation}
\bibinfo{author}{Thatcher, R.}, \bibinfo{author}{Biver, C.}, \bibinfo{author}{Gomez, J.}, \bibinfo{author}{North, D.}, \bibinfo{author}{Curtin, R.}, \bibinfo{author}{Walker, R.}, \bibinfo{author}{Salazar, A.}, \bibinfo{year}{2001}.
\newblock \bibinfo{title}{Estimation of the eeg power spectrum using mri t2 relaxation time in traumatic brain injury}.
\newblock \bibinfo{journal}{Clinical Neurophysiology} \bibinfo{volume}{112}, \bibinfo{pages}{1729--1745}.
\bibitem[{Thatcher et~al.(2016)Thatcher, Palmero-Soler, North and Biver}]{thatcher2016intelligence}
\bibinfo{author}{Thatcher, R.}, \bibinfo{author}{Palmero-Soler, E.}, \bibinfo{author}{North, D.}, \bibinfo{author}{Biver, C.}, \bibinfo{year}{2016}.
\newblock \bibinfo{title}{Intelligence and eeg measures of information flow: efficiency and homeostatic neuroplasticity}.
\newblock \bibinfo{journal}{Scientific reports} \bibinfo{volume}{6}, \bibinfo{pages}{1--10}.
\bibitem[{Thatcher and Lubar(2009)}]{thatcher2009history}
\bibinfo{author}{Thatcher, R.W.}, \bibinfo{author}{Lubar, J.F.}, \bibinfo{year}{2009}.
\newblock \bibinfo{title}{History of the scientific standards of qeeg normative databases}.
\newblock \bibinfo{journal}{Introduction to quantitative EEG and neurofeedback: Advanced theory and applications} \bibinfo{volume}{2}, \bibinfo{pages}{29--59}.
\bibitem[{Thatcher et~al.(1989)Thatcher, Walker, Gerson and Geisler}]{thatcher1989eeg}
\bibinfo{author}{Thatcher, R.W.}, \bibinfo{author}{Walker, R.}, \bibinfo{author}{Gerson, I.}, \bibinfo{author}{Geisler, F.}, \bibinfo{year}{1989}.
\newblock \bibinfo{title}{Eeg discriminant analyses of mild head trauma}.
\newblock \bibinfo{journal}{Electroencephalography and clinical neurophysiology} \bibinfo{volume}{73}, \bibinfo{pages}{94--106}.
\bibitem[{Van~Steenkiste et~al.(2020)Van~Steenkiste, van Loon and Crevecoeur}]{van2020transfer}
\bibinfo{author}{Van~Steenkiste, G.}, \bibinfo{author}{van Loon, G.}, \bibinfo{author}{Crevecoeur, G.}, \bibinfo{year}{2020}.
\newblock \bibinfo{title}{Transfer learning in ecg classification from human to horse using a novel parallel neural network architecture}.
\newblock \bibinfo{journal}{Scientific Reports} \bibinfo{volume}{10}, \bibinfo{pages}{1--12}.
\bibitem[{Vishwanath(2021)}]{vishwanath2021detection}
\bibinfo{author}{Vishwanath, M.}, \bibinfo{year}{2021}.
\newblock \bibinfo{title}{Detection of Traumatic Brain Injury Using a Standard Machine Learning Pipeline in Mouse and Human Sleep Electroencephalogram}.
\newblock \bibinfo{publisher}{University of California, Irvine}.
\bibitem[{Vishwanath et~al.(2022)Vishwanath, Dutt, Rahmani, Lim and Cao}]{vishwanath2022label}
\bibinfo{author}{Vishwanath, M.}, \bibinfo{author}{Dutt, N.}, \bibinfo{author}{Rahmani, A.M.}, \bibinfo{author}{Lim, M.M.}, \bibinfo{author}{Cao, H.}, \bibinfo{year}{2022}.
\newblock \bibinfo{title}{Label alignment improves eeg-based machine learning-based classification of traumatic brain injury}, in: \bibinfo{booktitle}{2022 44th Annual International Conference of the IEEE Engineering in Medicine \& Biology Society (EMBC)}, \bibinfo{organization}{IEEE}. pp. \bibinfo{pages}{3546--3549}.
\bibitem[{Vishwanath et~al.(2021)Vishwanath, Jafarlou, Shin, Dutt, Rahmani, Jones, Lim and Cao}]{vishwanath2021investigation}
\bibinfo{author}{Vishwanath, M.}, \bibinfo{author}{Jafarlou, S.}, \bibinfo{author}{Shin, I.}, \bibinfo{author}{Dutt, N.}, \bibinfo{author}{Rahmani, A.M.}, \bibinfo{author}{Jones, C.E.}, \bibinfo{author}{Lim, M.M.}, \bibinfo{author}{Cao, H.}, \bibinfo{year}{2021}.
\newblock \bibinfo{title}{Investigation of machine learning and deep learning approaches for detection of mild traumatic brain injury from human sleep electroencephalogram}, in: \bibinfo{booktitle}{2021 43rd Annual International Conference of the IEEE Engineering in Medicine \& Biology Society (EMBC)}, \bibinfo{organization}{IEEE}. pp. \bibinfo{pages}{6134--6137}.
\bibitem[{Vishwanath et~al.(2020a)Vishwanath, Jafarlou, Shin, Dutt, Rahmani, Lim and Cao}]{vishwanath2020classification}
\bibinfo{author}{Vishwanath, M.}, \bibinfo{author}{Jafarlou, S.}, \bibinfo{author}{Shin, I.}, \bibinfo{author}{Dutt, N.}, \bibinfo{author}{Rahmani, A.M.}, \bibinfo{author}{Lim, M.M.}, \bibinfo{author}{Cao, H.}, \bibinfo{year}{2020}a.
\newblock \bibinfo{title}{Classification of electroencephalogram in a mouse model of traumatic brain injury using machine learning approaches}, in: \bibinfo{booktitle}{2020 42nd Annual International Conference of the IEEE Engineering in Medicine \& Biology Society (EMBC)}, \bibinfo{organization}{IEEE}. pp. \bibinfo{pages}{3335--3338}.
\bibitem[{Vishwanath et~al.(2020b)Vishwanath, Jafarlou, Shin, Lim, Dutt, Rahmani and Cao}]{vishwanath2020investigation}
\bibinfo{author}{Vishwanath, M.}, \bibinfo{author}{Jafarlou, S.}, \bibinfo{author}{Shin, I.}, \bibinfo{author}{Lim, M.M.}, \bibinfo{author}{Dutt, N.}, \bibinfo{author}{Rahmani, A.M.}, \bibinfo{author}{Cao, H.}, \bibinfo{year}{2020}b.
\newblock \bibinfo{title}{Investigation of machine learning approaches for traumatic brain injury classification via eeg assessment in mice}.
\newblock \bibinfo{journal}{Sensors} \bibinfo{volume}{20}, \bibinfo{pages}{2027}.
\bibitem[{Vo et~al.(2022)Vo, Vishwanath, Srinivasan, Dutt and Cao}]{vo2022composing}
\bibinfo{author}{Vo, K.}, \bibinfo{author}{Vishwanath, M.}, \bibinfo{author}{Srinivasan, R.}, \bibinfo{author}{Dutt, N.}, \bibinfo{author}{Cao, H.}, \bibinfo{year}{2022}.
\newblock \bibinfo{title}{Composing graphical models with generative adversarial networks for eeg signal modeling}, in: \bibinfo{booktitle}{ICASSP 2022-2022 IEEE International Conference on Acoustics, Speech and Signal Processing (ICASSP)}, \bibinfo{organization}{IEEE}. pp. \bibinfo{pages}{1231--1235}.
\bibitem[{Voulodimos et~al.(2018)Voulodimos, Doulamis, Doulamis, Protopapadakis et~al.}]{voulodimos2018deep}
\bibinfo{author}{Voulodimos, A.}, \bibinfo{author}{Doulamis, N.}, \bibinfo{author}{Doulamis, A.}, \bibinfo{author}{Protopapadakis, E.}, et~al., \bibinfo{year}{2018}.
\newblock \bibinfo{title}{Deep learning for computer vision: A brief review}.
\newblock \bibinfo{journal}{Computational intelligence and neuroscience} \bibinfo{volume}{2018}.
\bibitem[{Wan et~al.(2021)Wan, Yang, Huang, Zeng and Liu}]{wan2021review}
\bibinfo{author}{Wan, Z.}, \bibinfo{author}{Yang, R.}, \bibinfo{author}{Huang, M.}, \bibinfo{author}{Zeng, N.}, \bibinfo{author}{Liu, X.}, \bibinfo{year}{2021}.
\newblock \bibinfo{title}{A review on transfer learning in eeg signal analysis}.
\newblock \bibinfo{journal}{Neurocomputing} \bibinfo{volume}{421}, \bibinfo{pages}{1--14}.
\bibitem[{Weiss and Mueller(2003)}]{weiss2003contribution}
\bibinfo{author}{Weiss, S.}, \bibinfo{author}{Mueller, H.M.}, \bibinfo{year}{2003}.
\newblock \bibinfo{title}{The contribution of eeg coherence to the investigation of language}.
\newblock \bibinfo{journal}{Brain and language} \bibinfo{volume}{85}, \bibinfo{pages}{325--343}.
\bibitem[{Wen et~al.(2020)Wen, Sun, Yang, Song, Gao, Wang and Xu}]{wen2020time}
\bibinfo{author}{Wen, Q.}, \bibinfo{author}{Sun, L.}, \bibinfo{author}{Yang, F.}, \bibinfo{author}{Song, X.}, \bibinfo{author}{Gao, J.}, \bibinfo{author}{Wang, X.}, \bibinfo{author}{Xu, H.}, \bibinfo{year}{2020}.
\newblock \bibinfo{title}{Time series data augmentation for deep learning: A survey}.
\newblock \bibinfo{journal}{arXiv preprint arXiv:2002.12478} .
\bibitem[{Widmann et~al.(2015)Widmann, Schr{\"o}ger and Maess}]{widmann2015digital}
\bibinfo{author}{Widmann, A.}, \bibinfo{author}{Schr{\"o}ger, E.}, \bibinfo{author}{Maess, B.}, \bibinfo{year}{2015}.
\newblock \bibinfo{title}{Digital filter design for electrophysiological data--a practical approach}.
\newblock \bibinfo{journal}{Journal of neuroscience methods} \bibinfo{volume}{250}, \bibinfo{pages}{34--46}.
\bibitem[{Willie et~al.(2012)Willie, Lim, Bennett, Azarion, Schwetye and Brody}]{willie2012controlled}
\bibinfo{author}{Willie, J.T.}, \bibinfo{author}{Lim, M.M.}, \bibinfo{author}{Bennett, R.E.}, \bibinfo{author}{Azarion, A.A.}, \bibinfo{author}{Schwetye, K.E.}, \bibinfo{author}{Brody, D.L.}, \bibinfo{year}{2012}.
\newblock \bibinfo{title}{Controlled cortical impact traumatic brain injury acutely disrupts wakefulness and extracellular orexin dynamics as determined by intracerebral microdialysis in mice}.
\newblock \bibinfo{journal}{Journal of neurotrauma} \bibinfo{volume}{29}, \bibinfo{pages}{1908--1921}.
\bibitem[{Wu et~al.(2015)Wu, Quinlan, Dodakian, McKenzie, Kathuria, Zhou, Augsburger, See, Le, Srinivasan et~al.}]{wu2015connectivity}
\bibinfo{author}{Wu, J.}, \bibinfo{author}{Quinlan, E.B.}, \bibinfo{author}{Dodakian, L.}, \bibinfo{author}{McKenzie, A.}, \bibinfo{author}{Kathuria, N.}, \bibinfo{author}{Zhou, R.J.}, \bibinfo{author}{Augsburger, R.}, \bibinfo{author}{See, J.}, \bibinfo{author}{Le, V.H.}, \bibinfo{author}{Srinivasan, R.}, et~al., \bibinfo{year}{2015}.
\newblock \bibinfo{title}{Connectivity measures are robust biomarkers of cortical function and plasticity after stroke}.
\newblock \bibinfo{journal}{Brain} \bibinfo{volume}{138}, \bibinfo{pages}{2359--2369}.
\bibitem[{Xu et~al.(2020)Xu, Xu, Ke, An, Liu and Ming}]{xu2020cross}
\bibinfo{author}{Xu, L.}, \bibinfo{author}{Xu, M.}, \bibinfo{author}{Ke, Y.}, \bibinfo{author}{An, X.}, \bibinfo{author}{Liu, S.}, \bibinfo{author}{Ming, D.}, \bibinfo{year}{2020}.
\newblock \bibinfo{title}{Cross-dataset variability problem in eeg decoding with deep learning}.
\newblock \bibinfo{journal}{Frontiers in human neuroscience} \bibinfo{volume}{14}, \bibinfo{pages}{103}.
\bibitem[{Yael et~al.(2018)Yael, Vecht and Bar-Gad}]{yael2018filter}
\bibinfo{author}{Yael, D.}, \bibinfo{author}{Vecht, J.J.}, \bibinfo{author}{Bar-Gad, I.}, \bibinfo{year}{2018}.
\newblock \bibinfo{title}{Filter-based phase shifts distort neuronal timing information}.
\newblock \bibinfo{journal}{Eneuro} \bibinfo{volume}{5}.
\bibitem[{Zhang et~al.(2019)Zhang, Wang, Yue, Zhang, Peng and Hu}]{zhang2019cross}
\bibinfo{author}{Zhang, F.}, \bibinfo{author}{Wang, F.}, \bibinfo{author}{Yue, L.}, \bibinfo{author}{Zhang, H.}, \bibinfo{author}{Peng, W.}, \bibinfo{author}{Hu, L.}, \bibinfo{year}{2019}.
\newblock \bibinfo{title}{Cross-species investigation on resting state electroencephalogram}.
\newblock \bibinfo{journal}{Brain Topography} \bibinfo{volume}{32}, \bibinfo{pages}{808--824}.
\bibitem[{Zhang et~al.(2021)Zhang, Robinson, Lee and Guan}]{zhang2021adaptive}
\bibinfo{author}{Zhang, K.}, \bibinfo{author}{Robinson, N.}, \bibinfo{author}{Lee, S.W.}, \bibinfo{author}{Guan, C.}, \bibinfo{year}{2021}.
\newblock \bibinfo{title}{Adaptive transfer learning for eeg motor imagery classification with deep convolutional neural network}.
\newblock \bibinfo{journal}{Neural Networks} \bibinfo{volume}{136}, \bibinfo{pages}{1--10}.

\end{thebibliography}






\end{document}